\documentclass[journal]{IEEEtran}
\usepackage{graphicx}
\usepackage{comment}
\usepackage{amsmath,amssymb} 
\usepackage{color}

\usepackage{footnote}
\usepackage{floatrow}
\usepackage{dirtytalk}
\usepackage{boldline}
\usepackage{multirow}
\usepackage{rotating}

\usepackage{subcaption}
\usepackage{color,soul}
\usepackage{multirow}
\usepackage{dirtytalk}
\usepackage{xcolor}
\usepackage{tabularx,booktabs}
\newcolumntype{Y}{>{\centering\arraybackslash}X}

\newcommand{\eg}{{\it e.g. }}

\newcommand{\bfsection}[1]{\vspace*{0.1cm}\noindent\textbf{#1.}}
\usepackage[pagebackref=true,breaklinks=true,letterpaper=true,colorlinks,bookmarks=false]{hyperref}

\begin{document}
%
\title{A Domain Gap Aware Generative Adversarial Network for Multi-domain Image Translation}

%
%
%

	\author{{Wenju~Xu},
		and~Guanghui~Wang,~\IEEEmembership{Senior Member,~IEEE}
		\thanks{W. Xu is with the OPPO US Research Center, InnoPeak Technology Inc, Palo Alto, CA, USA. Email: xuwenju123@gmail.com.}
		\thanks{G. Wang is with the Department of Computer Science, Ryerson University, 350 Victoria St, Toronto, ON M5B 2K3. Email: wangcs@ryerson.ca}.
\thanks{This work was partly supported by the Natural Sciences and Engineering Research Council of Canada (NSERC) under grant no. RGPIN-2021-04244, and the United States Department of Agriculture (USDA) under Grant no. 2019-67021-28996.}}

\maketitle

\begin{abstract}
Recent image-to-image translation models have shown great success in mapping local textures between two domains. Existing approaches rely on a cycle-consistency constraint that supervises the generators to learn an inverse mapping. However, learning the inverse mapping introduces extra trainable parameters and it is unable to learn the inverse mapping for some domains. As a result, they are ineffective in the scenarios where (i) multiple visual image domains are involved; (ii) both structure and texture transformations are required; and (iii) semantic consistency is preserved. To solve these challenges, the paper proposes a unified model to translate images across multiple domains with significant domain gaps. Unlike previous models that constrain the generators with the ubiquitous cycle-consistency constraint to achieve the content similarity, the proposed model employs a perceptual self-regularization constraint. With a single unified generator, the model can maintain consistency over the global shapes as well as the local texture information across multiple domains. Extensive qualitative and quantitative evaluations demonstrate the effectiveness and superior performance over state-of-the-art models. It is more effective in representing shape deformation in challenging mappings with significant dataset variation across multiple domains.
\end{abstract}


%
\IEEEpeerreviewmaketitle

\section{Introduction}
Image-to-image translation is to map input images from a source domain to a target domain through the domain-specific characteristic transformation, including inpainting \cite{chen2018high}, super-resolution \cite{johnson2016perceptual}, colorization \cite{zhang2016colorful}, etc. The training images are collected separately in terms of domain, and the use of discriminator criticizing domain-specific characteristics provides domain-level supervision for the generator training. Combined with other regularization, the generator learns a translation mapping between the source domain and the target domain. 
  
\begin{figure}[t]
\begin{center}
\includegraphics[width=0.96\textwidth]{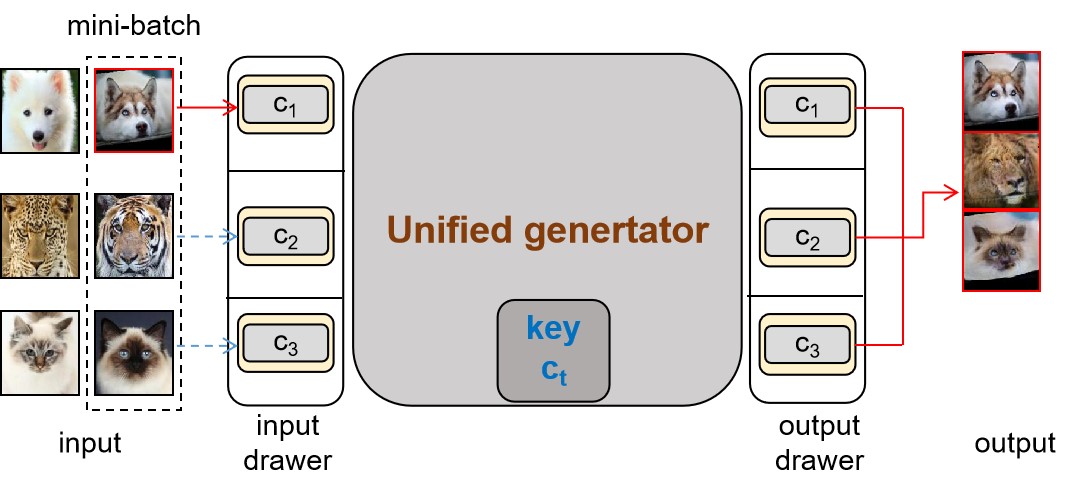}
\end{center}
\vspace{-0.6cm} 
\caption{\footnotesize Introduction of the proposed model. It works as a closet and realizes multiple-domain translation by stacking the drawers. Each input drawer corresponds to one input domain and each output drawer corresponds to one output domain. Users drop observed images into the input drawers, the unified generator can generate the translated images in the output drawers.
}\label{intro}

\end{figure}

\begin{figure*}[th]
\begin{center}
    \centering
    \includegraphics[width=1\textwidth]{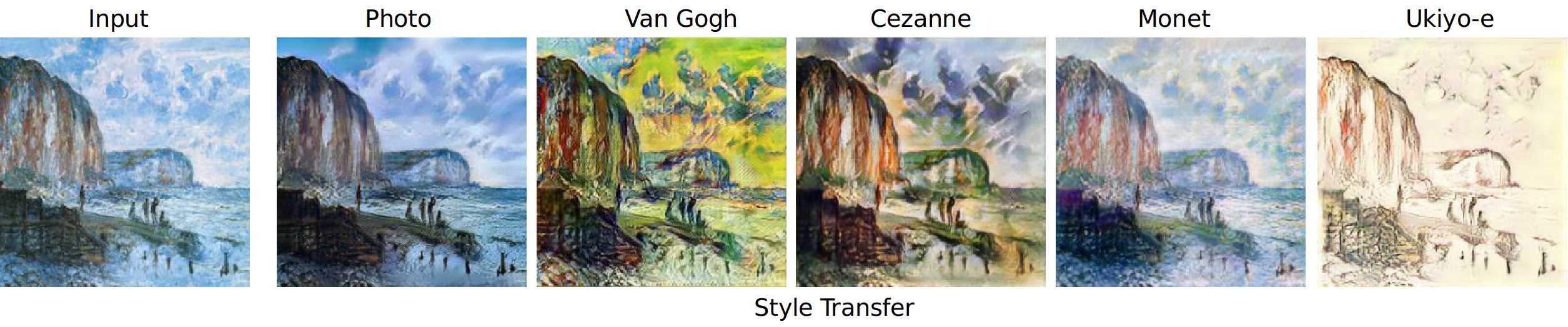}
    \includegraphics[width=0.98\textwidth]{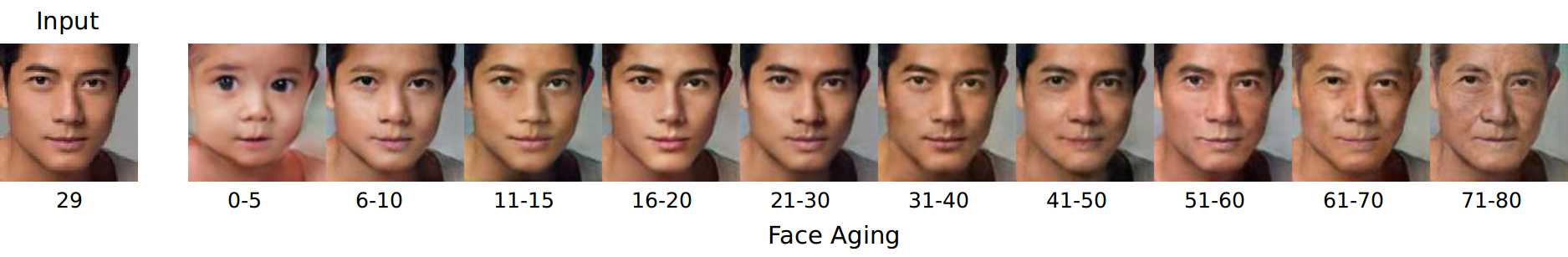}
    \includegraphics[width=0.98\textwidth]{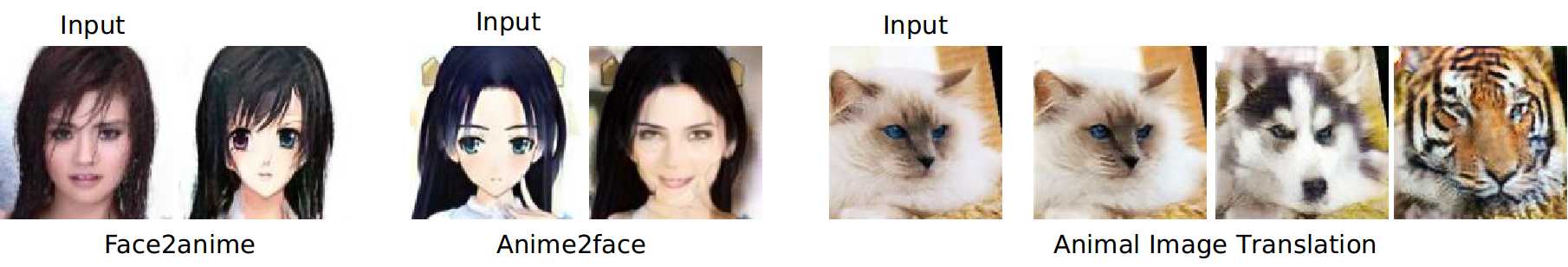}
\end{center}
\vspace{-0.6cm} 
\caption{
Some examples of the translated images by the proposed model across multiple domains with large domain gaps. ({\it top}) Collection image transfer. ({\it center}) Face aging. ({\it bottom}) face transformation.}\label{Demo}
\end{figure*} 
  
In recent years, there is a trend to make a trade-off between structure preservation and shape deformation in image translation tasks. One approach is to maintain the global structure and slightly change local textures \cite{pix2pix2016}, such as face attributes changing \cite{shen2017learning}. Another approach is to exploit shape deformation according to high-level semantic constraint \cite{liu2019few}, like cats to dogs. However, there exist various mappings between each domain pair given different regularization and training schemes. How to make a balance between these two requirements is still an open problem. 

The CycleGAN \cite{CycleGAN2017} has shown impressive results on patch-level image translation between two domains, such as translating stripes between zebras and horses. It introduces the most common regularization as the cycle-consistency restraint that learns a reverse mapping for additional supervision. Cycle consistency in the image space tends to preserve pixel-wise correspondence between the input and the output. It is helpful in tasks where the change is mostly on textures (e.g., horse to zebra) but recent findings \cite{yang2018esther,wu2019transgaga} have shown that being able to invert a mapping does not necessarily lead to successful translations for some domains, such as cats to dogs. It prevents the model from learning mappings that involve large shape differences. To promote shape deformation, previous works seek to increase the receptive field to discriminate on both the global and local features, match statistics of discriminatory features, penalizing distances in the latent space. However, this additional regularization introduces extra training parameters and computational costs, which are critical when dealing with multi-domains image translation.

In this work, we propose a unified model, named as UMIT, for multi-domain image translation. This model maintains consistency over both the global shape and the local texture in each domain by exploiting domain-specific information \cite{ruder2017overview}. We introduce the input and output drawers to preserve local textures. The working mechanism is illustrated in Figure \ref{intro}, where we put input images from one domain in the same drawer and generate the outputs from the corresponding output drawers. This mechanism facilitates disentangling generations in one framework. Furthermore, we efficiently integrate the discriminator with a multi-scale classifier and dilated convolutional layers. With a small portion of additional parameters, it significantly increases the receptive field for low-frequent shape change. Instead of exploring the cycle-consistency, we take a perceptual loss to minimize the distance between the input image and the translated image and expect the output is visually similar to the input in terms of foreground and background. Based on this observation, our model proposes to utilize a single generator that is trained with a self-regularization term to enforce perceptual similarity between the output and the input, making the training process simpler and more efficient. 

Extensive experiments demonstrate that the proposed model works surprisingly well in a wide variety of image translations involving multiple image domains, achieving superior qualitative and quantitative results as shown in Figure \ref{Demo}.  
We quantitatively and qualitatively demonstrate that our approach is more efficient and more effective by learning a unified model on challenging datasets, where the data comes from multiple resources and contains large shape deformations. We also carefully analyze the relation between our approach and CycleGAN and think that cycle-consistency is not a necessary assumption.

\begin{figure*}[t]
\centering
\includegraphics[width=0.45\linewidth]{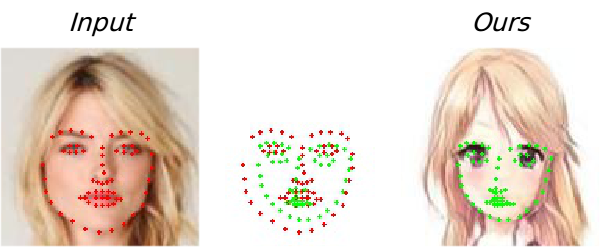}
\includegraphics[width=0.45\linewidth]{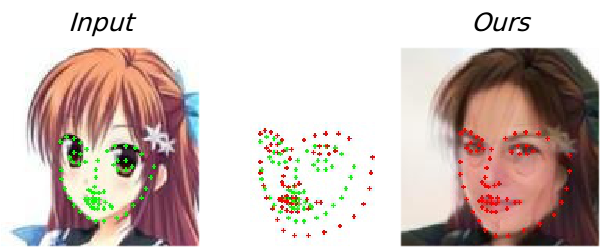} 
\caption{\footnotesize Landmarks on face images. Our approach translates the structure and texture appearance variations between the human and Anime domains (left: input; right: translation).}
\label{vis:faceLM}
\end{figure*}
\section{Related work}
Deep neural networks have shown great potential in dealing with real-world tasks \cite{zhou2014smart,xu2020adaptively, cen2019boosting, he2021sosd, XU2019195,xu2019adversarially, li2021colonoscopy}. Many deep learning based methods were proposed for image content understanding \cite{dong2021dual, Xu_2020_WACV} and image content generation tasks \cite{huang2020fx,XU2019570,dumoulin2016adversarially}. Especially, the content conditioned image synthesis task, which generates images based on the input content images via learning a mapping function between the input domain and target domain, has attracted a lot of attentions. It requires to not only create realistic synthesized images but also preserve the content structural similarity. Lots of works now are being introduced for this purpose. 

\bfsection{Generative Adversarial Network} Generative adversarial networks (GANs)~\cite{goodfellow2014generative} have shown impressive results in image generation~\cite{dumoulin2016adversarially,tolstikhin2017wasserstein}, image translation~\cite{zhu2017toward,larsen2016autoencoding,xu2021drb}, super-resolution imaging~\cite{johnson2016perceptual,ledig2016photo}, and face image synthesis~\cite{tran2017DRGAN,zhang2017age}. These GAN models are trained to minimize the discrepancy between distributions of the training data and unobserved generations by adversarially learning between two neural networks. WGAN \cite{arjovsky2017wasserstein} and WGAN-gp \cite{gulrajani2017improved} use W-distance to remedy the model collapse. Several recent efforts explore conditional GAN given in various image translation tasks to learn the mapping conditioned on prior constraints such as labels, text embedding, or images. Our work focuses on using a unified GAN model for multi-domain image translation. 

\bfsection{Image-to-Image Translation}
The goal of image-to-image translation is to learn a mapping between paired/unpaired image collections.
Pix2pix~\cite{pix2pix2016} handles this task by learning a one-directional mapping from the source to the target domain with paired images as supervised signals. But it is not suitable for tasks where paired images are not available. To alleviate the burden of obtaining data pairs, unsupervised image-to-image translation approaches have been proposed~\cite{CycleGAN2017,kim2017learning,liu2017unsupervised,yi2017dualgan}, which resort to the cycle consistency constraint for additional supervision. Similar ideas can also be found in~\cite{yi2017dualgan,taigman2016unsupervised,bousmalis2016unsupervised}. 

Recent multimodal image translation methods, such as BiCycleGAN~\cite{zhu2017toward}, MUNIT~ \cite{huang2018multimodal}, DRIT~\cite{lee2018diverse}, model a distribution of possible outputs by producing diverse translated samples instead of one deterministic sample for each single input image. This one-to-many mapping seeks to cover all possible mappings between two domains. However, few of them are effective to learn a map when the cross-domain gap (e.g., object shape) becomes large. The GANimorph~\cite{gokaslan2018improving} handles this task by utilizing dilated convolutions and multi-scale features in both discriminator and generator for larger receptive fields. WarpGAN~\cite{shi2019warpgan} manipulates the predicted control points to warp the photo into a caricature. Our model does not rely on cycle-consistency and seeks to bridge the domain gap by exploring multi-scale information (Figure \ref{vis:faceLM}).

Another type of approaches \cite{ma2017pose,cao2018carigans,CelebAMask-HQ} separately learns an appearance model and a structure model. The appearance model is to estimate the structure representation (landmarks or skeleton) of the object and synthesize a new sample-based on (manipulated) landmarks or skeleton. While the structure model is to manipulate the landmarks or skeleton based on pre-defined structural relation. In this sense, this type of approaches does not require the generative models to handle the shape deformation, but only synthesize samples conditioned on labels. Therefore, these approaches require image-label pairs to train both the appearance model and the structure model. TransGaGa~\cite{wu2019transgaga} proposes to learn the appearance and geometry latent space separately. However, the learned geometry latent representation is dominant, tending to generate new images purely conditioned on the geometry latent code without appearance preserved. In addition, this type of approaches deals with the geometry and appearance separately and fully converts the appearance of the source domain into that of the target domain without preserving the consistency between the input images and output images. However, our proposed model aims at learning a unified model in an unsupervised or self-supervised manner. Our task is more challenging in terms of \say{unsupervision}. In a direct way, we seek to promote the capacity of the generative model for shape transformation and consistency preservation. For illustration, the landmarks of both the human face and the generated Anime face are demonstrated in Figure \ref{vis:faceLM}. The face shapes are depicted by the landmarks. Our model does not rely on the landmarks but is able to transform the face and preserve the semantic consistency.

\bfsection{Multi-Domain Image Translation} This task is to translate images into different domains using a unified model. Unlike learning the mapping between two domains, it deals with the scalability of unsupervised image translation methods. Recent works have shown impressive results in multi-domain image-to-image translation using one single model~\cite{choi2018stargan,liu2018unified,liu2019stgan}. Based on CycleGAN, StarGAN~\cite{choi2018stargan} proposes a unified model that shares the generator and discriminator by all domains. Additionally, it integrates a classifier to predict domain probability. UFDN~\cite{liu2018unified} and GANimation~\cite{pumarola2018ganimation} learn a continuous latent manifold to perform a continuous cross-domain image translation and manipulation on human face dataset. However, these multi-domain models are more sensitive to the domain gap since training a classifier on the existing image translation dataset is challenging given the number of images. The ill-trained classifier fails to provide regularization when training the generator. To avoid this, ComboGAN~\cite{anoosheh2018combogan} and singleGAN~\cite{yu2018singlegan} employ multiple discriminators to train one unified generator, while they significantly increase the number of trainable parameters. The most recent FUNIT \cite{liu2019few} focuses on few-shot learning. Based on MUNIT, it proposes a unified discriminator that works well on a large number of domains with limited samples. Nevertheless, the performance of these methods is limited to the dataset that contains easy-recognized image domains.

\begin{figure*}[t]
\begin{subfigure}[b]{0.44\textwidth}
\includegraphics[width=0.98\textwidth]{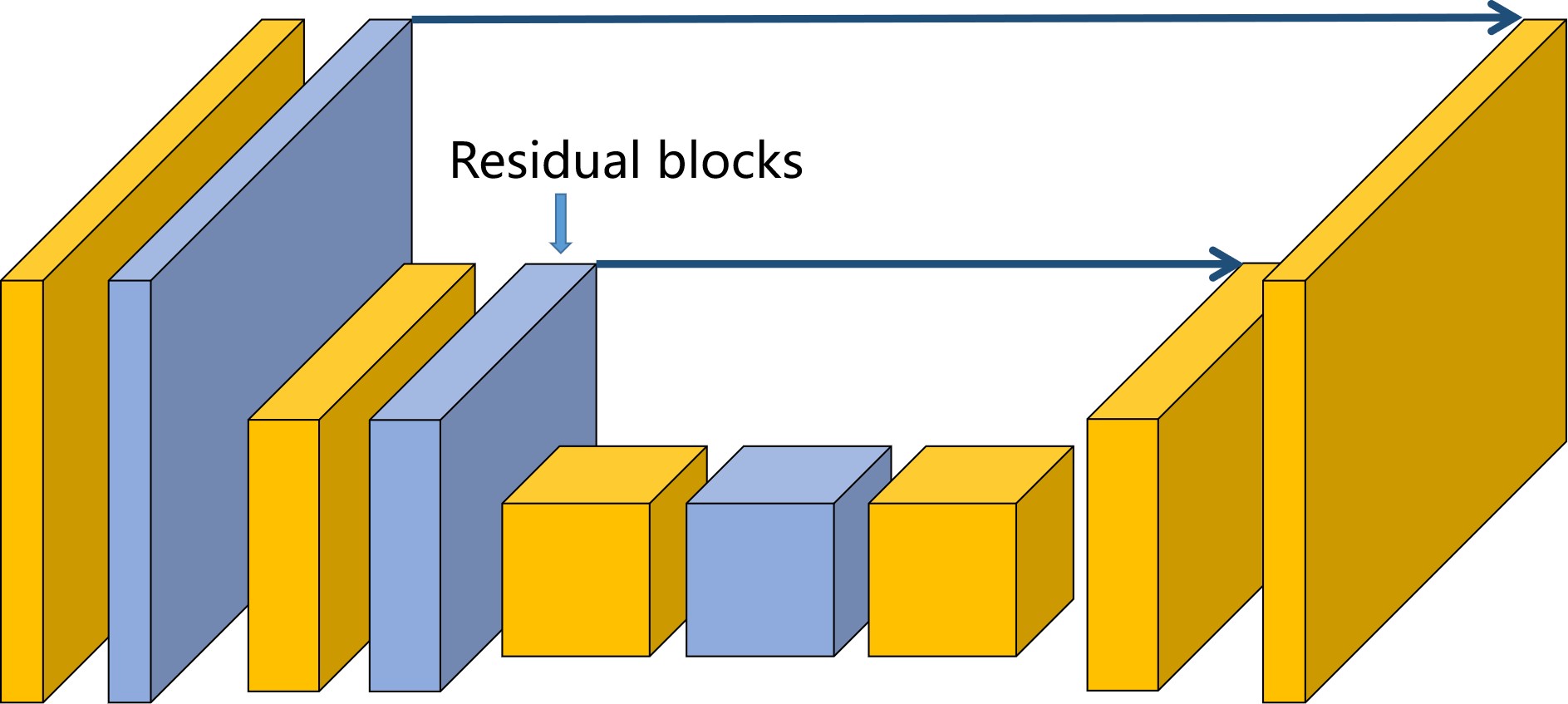}
\caption{}
\label{generator}
\end{subfigure}
\begin{subfigure}[b]{0.44\textwidth}
\includegraphics[width=0.98\textwidth]{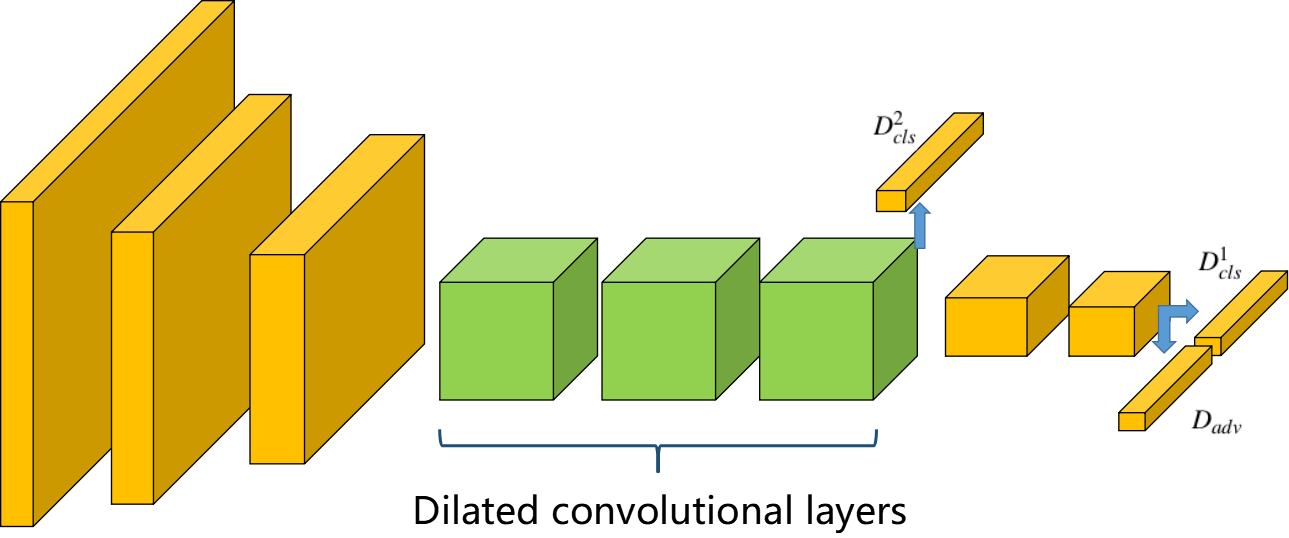}
\caption{}
\label{discriminator}
\end{subfigure}
\vspace{-0.4cm} 
\caption{\footnotesize (a) The generator is a combination of Unet and Resnet. The skip
connections and residual blocks are combined via concatenation as opposed to addition.
(b) The discriminator architecture is a fully-convolutional
network. Each colored block represents a convolutional layer; block labels indicate filter sizes. In addition to the global context from the dilations, two additional classifiers are employed to enforce the network's view of the domain-specific feature.
}
\label{architecture}
\end{figure*}

\begin{figure}[t]
\begin{center}
\includegraphics[width=0.98\textwidth]{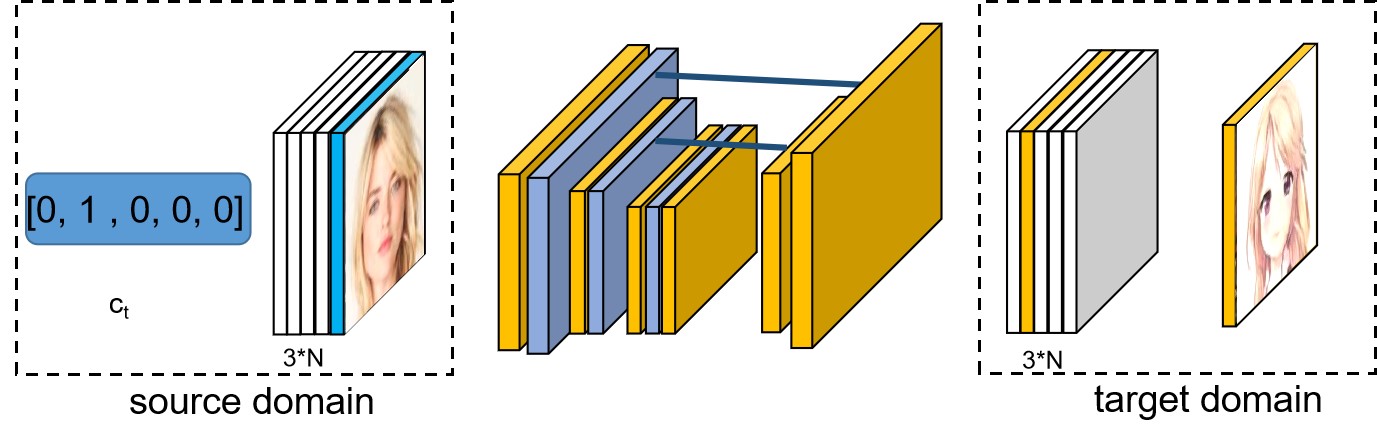}
\end{center}
\vspace{-0.3cm}  
\caption{\footnotesize The architecture of the input and output drawers.}
\label{overview}
\end{figure}

\section{Problem Formulation}
Given image sets $\{X_c\}^N_{c=1}$ from $N$ domains, our goal is to learn a pair of generator-discriminator $\{G-D\}$ that transforms domain-specific properties across all of the domains. Additionally, we integrate a multi-scale classifier into the discriminator with most parameters shared. Let us use subscripts $s$ and $t$ to indicate the source and target domain, and $c$ for domain label. This applies to represent transformation between any domain pair ({\it i.e.,} to map input image to one of the target domains and vice-versa). The problem is formulated as $G(x_s, c_t)\to x_t$. We will elaborate the proposed approach in this section.
 
\subsection{Network Architecture}
\bfsection{Unified Generator}
The architecture of our generator is illustrated in Figure \ref{generator}. It integrates residual blocks at multiple layers into a Unet, allowing the network to aggregate multi-scale information for the transformation of both higher- and lower-spatial resolution features. This structure takes both the images and the domain labels as inputs. In addition, we propose a novel input-output drawer mechanism, allowing the unified generator to disentangle local textures at pixel-level when all learnable parameters are shared.

\bfsection{Input-output Drawer}
The input-output drawer structure, as shown in Figure \ref{intro}, works as a domain container preserving domain-specific local textures. We stack the same number of drawers as involved domains for the input or output images. Let B, C, W, and H refer to the batch size, channel dimension, feature width, and height, respectively; $c_t$ indicates the target domains and $c_s$ represents the source domains. Given images $x_s \in \mathbb{R}^{B\times C\times H\times W}$ and the initial input drawer $T_{in}\in \mathbb{R}^{B\times [C\times N]\times H\times W}$, we assign each image separately to the input drawer $T_{in}$ as:
\begin{align}
\begin{aligned}
T_{in}[i,C\times c_s:C\times(c_s+1)] = x_s[i]
\end{aligned}
\end{align}
Then the input of generator is formed as a concatenation of $[T_{in}
, c_t] \in \mathbb{R}^{B \times(C \times N + c_t) \times H\times W}$. We perform a similar operation to receive output images from the output drawer given $T_{out} = G(T_{in}, c_t)$. The final images can be obtained by selecting the one indicated by the target domain label:
\begin{align}
\begin{aligned}
x_t[i] =  T_{out}[i,C\times c_t:C\times (c_t+1)]
\end{aligned}
\end{align}
where $i$ is the index used to iterate in the mini-batch and $C\times c_t$ is to select the target image channel-wisely. In this way, the generator can focus exclusively on the domain-specific feature translation (global shape information), leading to sharper and more realistic synthetic images. It does not need to learn separately a mapping for each domain pair. This process is depicted in Figure \ref{overview}.

\bfsection{Multi-scale Discriminator}
Using one unified model for Multi-domain image translation poses a significant challenge to the design of GAN discriminator. It requires to have a large receptive field for domain-specific global features. This would require a sophisticated architecture to capture the global features and differentiate the features according to the domain information. The patch discriminator has been proven to be effective for local texture translations. However, this patch-based view limits the network's awareness of global spatial information, which limits the generator's ability to perform coherent global shape changes. The dilated discriminator encourages global shape changes with a large receptive field. But these structures are not effective in the scenario of multi-domain translation. To alleviate these issues, we combine classifiers at multi-scales with the discriminator for domain-specific feature recognition, as shown in Figure \ref{discriminator}. This structure can provide globally consistent information for the generator training.

\subsection{Objective Function}
We aim at generating domain-consistent images with high visual quality in a unified model. We randomly assign the target domains for input images. The discriminator is to distinguish between the input images and the translated images, while the integrated classifier produces probability distributions over domain labels. We ignore the input-output drawer for expression simplicity as $G(x_s, c_t)\to x_t$. Our goal is to optimize the following objectives.

\bfsection{Adversarial Loss}
The original GAN matches two distributions of the real and fake images. However, the adversarial loss is potentially not continuous with respect to the parameters in the generator. This can locally saturate and lead to vanishing gradients in the discriminator. As a result, it is extremely difficult to train the GAN model. We take advantage of the Wasserstein GAN with gradient penalty (WGAN-GP) \cite{gulrajani2017improved,arjovsky2017wasserstein} to stabilize the training process. The adversarial loss is written as
\begin{align}
\begin{aligned}
\mathcal{L}_{adv} &= \mathbb{E}_{x\sim X}[D_{adv}(x)] + \mathbb{E}_{x\sim X}[-D_{adv}(G(x_s,c_t))]\\&+ 
\lambda_{gp}\mathbb{E}_{\hat{x}}[(||\triangledown_{\hat{x}}D_{adv}(\hat{x})||_2-1)^2]
\end{aligned}
\end{align}
where $\hat{x}$ is the linear interpolation between a real image and the generated one.  

\bfsection{Multi-scale Domain Classification Loss}
The results using $L_{adv}$ alone tend to have annoying artifacts, because the search for a local minimum of $L_{cls}$ may go through a path that resides outside the manifold of natural images. Thus, we combine $L_{cls}$ with $L_{adv}$ as the final objective function for $D$ to ensure that the search resides in that manifold and produce photo-realistic and domain-preserving images.
\begin{align}
\begin{aligned}
\mathcal{L}_{D} = \mathcal{L}_{adv} +  \sum_{i=1}^{2} \lambda_i \mathcal{L}_{cls}^i
\end{aligned}
\end{align}

We employ a typical softmax cross-entropy loss and regularize the outputs with the multi-class cross-entropy loss 
\begin{align}
\begin{aligned}
\mathcal{L}_{cls}^i& = \mathbb{E}_{x_s\sim X}[\log D_{cls}^i(c_s|x_s)+\log (D_{cls}^i(c_t|G(x_s,c_t))]
\end{aligned}
\end{align}
where the term $D_{cls}^i(x)$ represents a probability distribution over all domains. $\tilde{x}$ represents the generated images. Theoretically, minimizing $L_{cls}$ in the adversarial training scheme would 
learn a mapping $G$ that produces outputs consistent to the target domain. However, if the capacity is large enough, a network can map the input images to any randomly generated samples in the target domain. If the classifier is ill-trained, the network fails to map the input images to the target domain. Thus, other regularizations need to be introduced to ensure the learned function $G$ maps the input to the desired output.

\bfsection{ Perceptual Self-Regularization Loss}
To further constrain the learned mapping, the translated image should preserve the visual characteristics of the input image including color and pose, etc. In other words, the output and input need to share perceptual similarities. To this end, we impose a perceptual self-regularization constraint, which is modeled
by minimizing the distance between the input and output images. The perceptual loss is used for semantic consistency, which promotes the model to preserve the semantically related features, such as the color and the pose, while allowing us to change the domain-specific features. In contrast, the cycle loss seeks for pixel-level consistency. It either fails to maintain the semantic consistency or prevents the changes in the local patch.

We use a pre-trained VGG-19 network to compute style losses at multiple levels and one content loss in a way similar to the prior work. At the training stage, the target domain labels are randomly generated as $c_t = c_s[perm]$, where ${\it perm}$ stands for randomly generated permutation of data in the mini-batch. We attempt to minimize the perceptual loss of the translated image $x_t$ by considering the input source image $x_s$ as the content image while the $x_s[perm]$ as the style image. Similar to \cite{johnson2016perceptual,huang2017arbitrary}, we use the pre-trained VGG-19 to compute the loss.
\begin{align}
\label{eqx}
\begin{split}
\mathcal{L}_{vgg} = \mathcal{L}_{c} + \lambda \mathcal{L}_{s}
\end{split}
\end{align}
where the content loss is the Euclidean distance between the target features and the features in the output image.
\begin{align}
\label{eqx}
\begin{split}
\mathcal{L}_{c} =||\phi(x_t) - \phi(x_s)||
\end{split}
\end{align}
In our experiments, $\phi$ represents the features of $relu_{41}$ layers. Our style loss only matches the mean and standard deviation of the style features.
\begin{align}
\label{eqx}
\begin{split}
\mathcal{L}_{s} =\sum_{i=1}^L||\zeta(\psi_i(x_t)) - \zeta(\psi_i(x_s[perm]))||_2
\end{split}
\end{align}
where $\zeta$ is the {\it Gram matrix} and each $\psi_i$ denotes a layer in VGG-19 used to compute the style loss. In our experiments we use $relu_{21}$, $relu_{31}$, $relu_{41}$ layers with equal weights. 

\bfsection{Identity Preserving Loss}
As observed by previous works, the image translation GAN models lean to recognize the difference of color distributions instead of the texture/shape difference. To ensure the color consistency, we take the identity-preserving loss.
\begin{align}
\label{eqx}
\begin{split}
\mathcal{L}^{identity} &= \mathbb{E}_{x\sim X}||x-G(x,c_t)||_1
\end{split}
\end{align}

\bfsection{Final Objective Function} Through the above analysis, we train the proposed network by optimizing the following objective function.
\begin{align}
\begin{aligned}
\min_G \max_D \mathcal{L}_{D} + \lambda_{identity} \mathcal{L}^{identity}_{1} + \lambda_{vgg} \mathcal{L}_{vgg}
\end{aligned}
\end{align}

\begin{figure}[!h]
\centering
\includegraphics[width=0.45\linewidth]{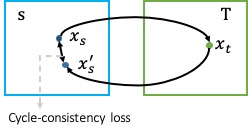}
\hspace{12pt}
\includegraphics[width=0.45\linewidth]{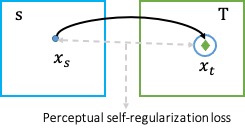} 
\caption{\footnotesize Comparison between the cycle consistency loss and perceptual self-Regularization loss. Given two image domains S and T represented by blue and green rectangles,
any point inside a rectangle represents a specific image in that domain. (Left): given a source image $x_s$, cycle-consistency loss requires the image translated back, $x_s^{\prime}$, to
be identical to the source image $x_s$. (Right): given a source image $x_s$, we synthesize images $x_t$ with shape variants (from a round dot to a rhombus) in the target domain. Instead of requiring the image to be translated back, we minimize the distance between $x_s$
and $x_t$, so that 
$x_t$ is encouraged to preserve the features of the original image $x_s$.}
\label{vis:remarks}
\end{figure}

{\noindent\bf Remarks}. To demonstrate the effect of different losses, we visualize the difference between the cycle loss and our perceptual self-Regularization loss in Figure \ref{vis:remarks}. The cycle loss seeks for pixel-level consistency. The main issue of cycle-consistency is that it assumes the translated images contain all the information of the input images in order to reconstruct the input images. This assumption leads to the preservation of source domain features, such as traces of objects and texture, resulting in unrealistic results. Moreover, the cycle-consistency constrains shape changes of source images. Our model proposes to use a single generator to learn directed mappings with a self-regularization term that enforces perceptual similarity between the output and the input, together with an adversarial term that enforces the output to appear like drawn from the target domain. The translated images preserve both high-level attributes and local texture from the target domain, relaxing the requirement of preserving source domain features for pixel-level consistency. 

\begin{figure*}[!t]
\centering
 \raisebox{0.005\textwidth }{\makebox[0.99\textwidth][l]{ \scriptsize \hspace{ 10pt}  Input \hspace{ 10pt}  Ours \hspace{ 10pt}  CycleGAN \hspace{ 10pt}  DRIT\hspace{ 10pt} StarGAN  \hspace{ 10pt}  MUNIT \hspace{ 10pt}FUNIT\hspace{ 24pt}Input\hspace{ 20pt}Ours\hspace{ 10pt}CycleGAN\hspace{ 12pt}DRIT\hspace{ 14pt}StarGAN\hspace{ 12pt}MUNIT\hspace{ 12pt}FUNIT}}
 \includegraphics[width=0.067\textwidth,height=0.067\textwidth]{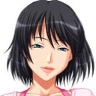}
  \hspace{-6pt}
 \includegraphics[width=0.067\textwidth,height=0.067\textwidth]{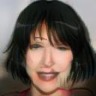}
  \hspace{-6pt}
 \includegraphics[width=0.067\textwidth,height=0.067\textwidth]{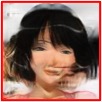}
  \hspace{-6pt}
 \includegraphics[width=0.067\textwidth,height=0.067\textwidth]{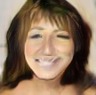}
  \hspace{-6pt}  
 \includegraphics[width=0.067\textwidth,height=0.067\textwidth]{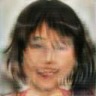}
  \hspace{-6pt}
  \includegraphics[width=0.067\textwidth,height=0.067\textwidth]{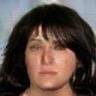}
   \hspace{-6pt}
   \includegraphics[width=0.067\textwidth,height=0.067\textwidth]{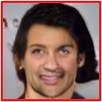}
  \hspace{-1pt}
 \includegraphics[width=0.067\textwidth,height=0.067\textwidth]{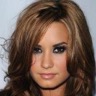}
  \hspace{-6pt}
 \includegraphics[width=0.067\textwidth,height=0.067\textwidth]{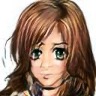}
  \hspace{-6pt}  
 \includegraphics[width=0.067\textwidth,height=0.067\textwidth]{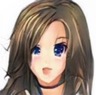}
  \hspace{-6pt}
  \includegraphics[width=0.067\textwidth,height=0.067\textwidth]{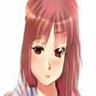}
  \hspace{-6pt}
 \includegraphics[width=0.067\textwidth,height=0.067\textwidth]{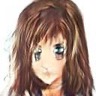}
  \hspace{-6pt}
 \includegraphics[width=0.067\textwidth,height=0.067\textwidth]{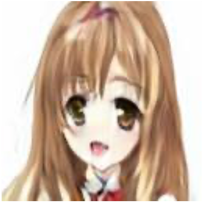}
  \hspace{-6pt}  
 \includegraphics[width=0.067\textwidth,height=0.067\textwidth]{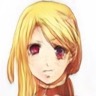}

\vspace{1pt}

 \includegraphics[width=0.067\textwidth,height=0.067\textwidth]{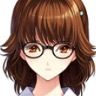}
  \hspace{-6pt}
 \includegraphics[width=0.067\textwidth,height=0.067\textwidth]{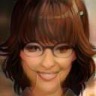}
  \hspace{-6pt}
 \includegraphics[width=0.067\textwidth,height=0.067\textwidth]{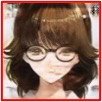}
  \hspace{-6pt}
 \includegraphics[width=0.067\textwidth,height=0.067\textwidth]{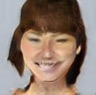}
  \hspace{-6pt}  
 \includegraphics[width=0.067\textwidth,height=0.067\textwidth]{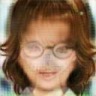}
  \hspace{-6pt}
  \includegraphics[width=0.067\textwidth,height=0.067\textwidth]{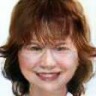}
   \hspace{-6pt}
   \includegraphics[width=0.067\textwidth,height=0.067\textwidth]{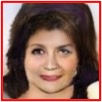}
  \hspace{-1pt}
 \includegraphics[width=0.067\textwidth,height=0.067\textwidth]{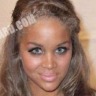}
  \hspace{-6pt}
 \includegraphics[width=0.067\textwidth,height=0.067\textwidth]{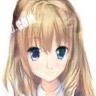}
  \hspace{-6pt}  
 \includegraphics[width=0.067\textwidth,height=0.067\textwidth]{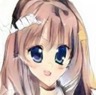}
  \hspace{-6pt}
  \includegraphics[width=0.067\textwidth,height=0.067\textwidth]{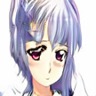}
  \hspace{-6pt}
 \includegraphics[width=0.067\textwidth,height=0.067\textwidth]{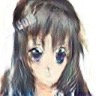}
  \hspace{-6pt}
 \includegraphics[width=0.067\textwidth,height=0.067\textwidth]{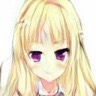}
  \hspace{-6pt}  
 \includegraphics[width=0.067\textwidth,height=0.067\textwidth]{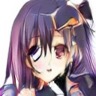}

\vspace{1pt}

 \includegraphics[width=0.067\textwidth,height=0.067\textwidth]{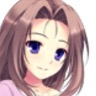}
  \hspace{-6pt}
 \includegraphics[width=0.067\textwidth,height=0.067\textwidth]{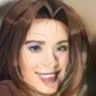}
  \hspace{-6pt}
 \includegraphics[width=0.067\textwidth,height=0.067\textwidth]{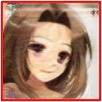}
  \hspace{-6pt}
 \includegraphics[width=0.067\textwidth,height=0.067\textwidth]{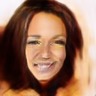}
  \hspace{-6pt}  
 \includegraphics[width=0.067\textwidth,height=0.067\textwidth]{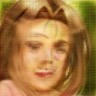}
  \hspace{-6pt}
  \includegraphics[width=0.067\textwidth,height=0.067\textwidth]{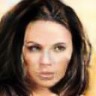}
   \hspace{-6pt}
   \includegraphics[width=0.067\textwidth,height=0.067\textwidth]{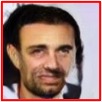}
  \hspace{-1pt}
 \includegraphics[width=0.067\textwidth,height=0.067\textwidth]{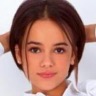}
  \hspace{-6pt}
 \includegraphics[width=0.067\textwidth,height=0.067\textwidth]{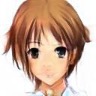}
  \hspace{-6pt}  
 \includegraphics[width=0.067\textwidth,height=0.067\textwidth]{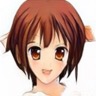}
  \hspace{-6pt}
  \includegraphics[width=0.067\textwidth,height=0.067\textwidth]{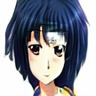}
  \hspace{-6pt}
 \includegraphics[width=0.067\textwidth,height=0.067\textwidth]{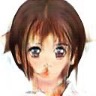}
  \hspace{-6pt}
 \includegraphics[width=0.067\textwidth,height=0.067\textwidth]{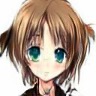}
  \hspace{-6pt}  
 \includegraphics[width=0.067\textwidth,height=0.067\textwidth]{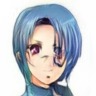}

\vspace{1pt}

 \includegraphics[width=0.067\textwidth,height=0.067\textwidth]{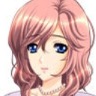}
  \hspace{-6pt}
 \includegraphics[width=0.067\textwidth,height=0.067\textwidth]{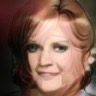}
  \hspace{-6pt}
 \includegraphics[width=0.067\textwidth,height=0.067\textwidth]{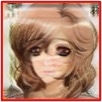}
  \hspace{-6pt}
 \includegraphics[width=0.067\textwidth,height=0.067\textwidth]{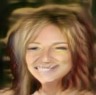}
  \hspace{-6pt}  
 \includegraphics[width=0.067\textwidth,height=0.067\textwidth]{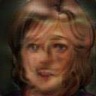}
  \hspace{-6pt}
  \includegraphics[width=0.067\textwidth,height=0.067\textwidth]{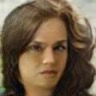}
   \hspace{-6pt}
   \includegraphics[width=0.067\textwidth,height=0.067\textwidth]{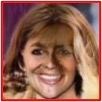}
  \hspace{-1pt}
 \includegraphics[width=0.067\textwidth,height=0.067\textwidth]{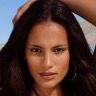}
  \hspace{-6pt}
 \includegraphics[width=0.067\textwidth,height=0.067\textwidth]{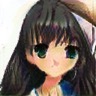}
  \hspace{-6pt}  
 \includegraphics[width=0.067\textwidth,height=0.067\textwidth]{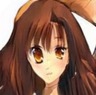}
  \hspace{-6pt}
  \includegraphics[width=0.067\textwidth,height=0.067\textwidth]{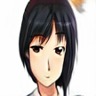}
  \hspace{-6pt}
 \includegraphics[width=0.067\textwidth,height=0.067\textwidth]{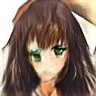}
  \hspace{-6pt}
 \includegraphics[width=0.067\textwidth,height=0.067\textwidth]{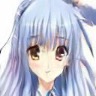}
  \hspace{-6pt}  
 \includegraphics[width=0.067\textwidth,height=0.067\textwidth]{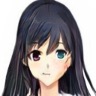}
 \raisebox{0.005\textwidth }{\makebox[0.99\textwidth][l]{ \footnotesize  \hspace{ 88pt} Anime to Human \hspace{ 200pt} Human to Anime}}

\caption{\footnotesize Visual comparisons on the human face $\leftrightarrow$ Anime face task.}\label{fig:Anime}
\vspace{-0.5cm} 
\end{figure*}

\begin{figure*}[!t]
\centering
 \raisebox{0.005\textwidth }{\makebox[0.99\textwidth][l]{ \scriptsize \hspace{ 10pt}  Input \hspace{ 10pt}  Ours \hspace{ 10pt}  CycleGAN \hspace{ 10pt}  DRIT\hspace{ 10pt} StarGAN  \hspace{ 10pt}  MUNIT \hspace{ 10pt}FUNIT\hspace{ 24pt}Input\hspace{ 20pt}Ours\hspace{ 10pt}CycleGAN\hspace{ 12pt}DRIT\hspace{ 14pt}StarGAN\hspace{ 12pt}MUNIT\hspace{ 12pt}FUNIT}}
 \includegraphics[width=0.067\textwidth,height=0.067\textwidth]{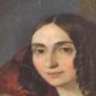}
  \hspace{-6pt}
 \includegraphics[width=0.067\textwidth,height=0.067\textwidth]{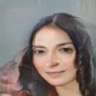}
  \hspace{-6pt}
 \includegraphics[width=0.067\textwidth,height=0.067\textwidth]{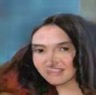}
  \hspace{-6pt}
 \includegraphics[width=0.067\textwidth,height=0.067\textwidth]{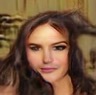}
  \hspace{-6pt}  
 \includegraphics[width=0.067\textwidth,height=0.067\textwidth]{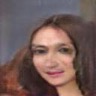}
  \hspace{-6pt}
  \includegraphics[width=0.067\textwidth,height=0.067\textwidth]{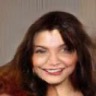}
   \hspace{-6pt}
   \includegraphics[width=0.067\textwidth,height=0.067\textwidth]{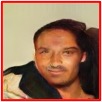}
  \hspace{-1pt}
 \includegraphics[width=0.067\textwidth,height=0.067\textwidth]{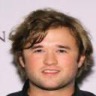}
  \hspace{-6pt}
 \includegraphics[width=0.067\textwidth,height=0.067\textwidth]{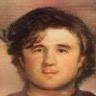}
  \hspace{-6pt}  
 \includegraphics[width=0.067\textwidth,height=0.067\textwidth]{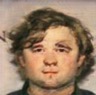}
  \hspace{-6pt}
  \includegraphics[width=0.067\textwidth,height=0.067\textwidth]{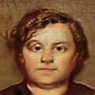}
  \hspace{-6pt}
 \includegraphics[width=0.067\textwidth,height=0.067\textwidth]{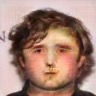}
  \hspace{-6pt}
 \includegraphics[width=0.067\textwidth,height=0.067\textwidth]{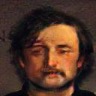}
  \hspace{-6pt}  
 \includegraphics[width=0.067\textwidth,height=0.067\textwidth]{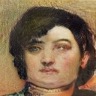}

\vspace{1pt}

 \includegraphics[width=0.067\textwidth,height=0.067\textwidth]{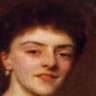}
  \hspace{-6pt}
 \includegraphics[width=0.067\textwidth,height=0.067\textwidth]{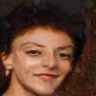}
  \hspace{-6pt}
 \includegraphics[width=0.067\textwidth,height=0.067\textwidth]{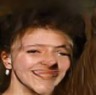}
  \hspace{-6pt}
 \includegraphics[width=0.067\textwidth,height=0.067\textwidth]{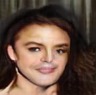}
  \hspace{-6pt}  
 \includegraphics[width=0.067\textwidth,height=0.067\textwidth]{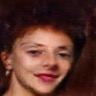}
  \hspace{-6pt}
  \includegraphics[width=0.067\textwidth,height=0.067\textwidth]{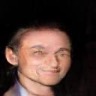}
   \hspace{-6pt}
   \includegraphics[width=0.067\textwidth,height=0.067\textwidth]{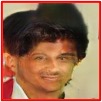}
  \hspace{-1pt}
 \includegraphics[width=0.067\textwidth,height=0.067\textwidth]{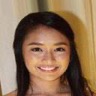}
  \hspace{-6pt}
 \includegraphics[width=0.067\textwidth,height=0.067\textwidth]{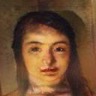}
  \hspace{-6pt}  
 \includegraphics[width=0.067\textwidth,height=0.067\textwidth]{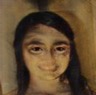}
  \hspace{-6pt}
  \includegraphics[width=0.067\textwidth,height=0.067\textwidth]{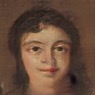}
  \hspace{-6pt}
 \includegraphics[width=0.067\textwidth,height=0.067\textwidth]{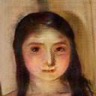}
  \hspace{-6pt}
 \includegraphics[width=0.067\textwidth,height=0.067\textwidth]{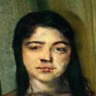}
  \hspace{-6pt}  
 \includegraphics[width=0.067\textwidth,height=0.067\textwidth]{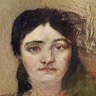}

\vspace{1pt}

 \includegraphics[width=0.067\textwidth,height=0.067\textwidth]{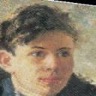}
  \hspace{-6pt}
 \includegraphics[width=0.067\textwidth,height=0.067\textwidth]{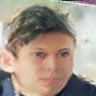}
  \hspace{-6pt}
 \includegraphics[width=0.067\textwidth,height=0.067\textwidth]{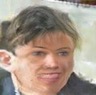}
  \hspace{-6pt}
 \includegraphics[width=0.067\textwidth,height=0.067\textwidth]{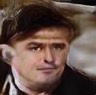}
  \hspace{-6pt}  
 \includegraphics[width=0.067\textwidth,height=0.067\textwidth]{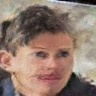}
  \hspace{-6pt}
  \includegraphics[width=0.067\textwidth,height=0.067\textwidth]{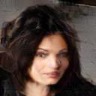}
   \hspace{-6pt}
   \includegraphics[width=0.067\textwidth,height=0.067\textwidth]{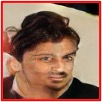}
  \hspace{-1pt}
 \includegraphics[width=0.067\textwidth,height=0.067\textwidth]{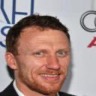}
  \hspace{-6pt}
 \includegraphics[width=0.067\textwidth,height=0.067\textwidth]{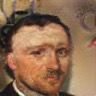}
  \hspace{-6pt}  
 \includegraphics[width=0.067\textwidth,height=0.067\textwidth]{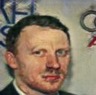}
  \hspace{-6pt}
  \includegraphics[width=0.067\textwidth,height=0.067\textwidth]{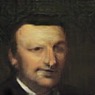}
  \hspace{-6pt}
 \includegraphics[width=0.067\textwidth,height=0.067\textwidth]{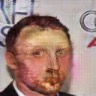}
  \hspace{-6pt}
 \includegraphics[width=0.067\textwidth,height=0.067\textwidth]{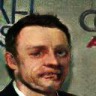}
  \hspace{-6pt}  
 \includegraphics[width=0.067\textwidth,height=0.067\textwidth]{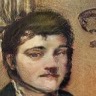}

\vspace{1pt}

 \includegraphics[width=0.067\textwidth,height=0.067\textwidth]{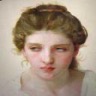}
  \hspace{-6pt}
 \includegraphics[width=0.067\textwidth,height=0.067\textwidth]{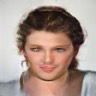}
  \hspace{-6pt}
 \includegraphics[width=0.067\textwidth,height=0.067\textwidth]{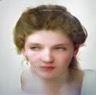}
  \hspace{-6pt}
 \includegraphics[width=0.067\textwidth,height=0.067\textwidth]{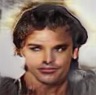}
  \hspace{-6pt}  
 \includegraphics[width=0.067\textwidth,height=0.067\textwidth]{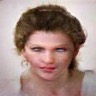}
  \hspace{-6pt}
  \includegraphics[width=0.067\textwidth,height=0.067\textwidth]{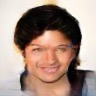}
   \hspace{-6pt}
   \includegraphics[width=0.067\textwidth,height=0.067\textwidth]{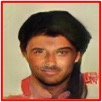}
  \hspace{-1pt}
 \includegraphics[width=0.067\textwidth,height=0.067\textwidth]{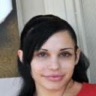}
  \hspace{-6pt}
 \includegraphics[width=0.067\textwidth,height=0.067\textwidth]{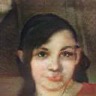}
  \hspace{-6pt}  
 \includegraphics[width=0.067\textwidth,height=0.067\textwidth]{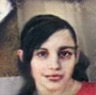}
  \hspace{-6pt}
  \includegraphics[width=0.067\textwidth,height=0.067\textwidth]{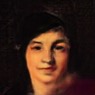}
  \hspace{-6pt}
 \includegraphics[width=0.067\textwidth,height=0.067\textwidth]{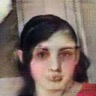}
  \hspace{-6pt}
 \includegraphics[width=0.067\textwidth,height=0.067\textwidth]{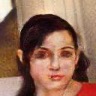}
  \hspace{-6pt}  
 \includegraphics[width=0.067\textwidth,height=0.067\textwidth]{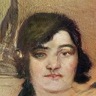}
 \raisebox{0.005\textwidth }{\makebox[0.99\textwidth][l]{ \footnotesize  \hspace{ 88pt} Portrait to Human \hspace{ 200pt} Human to Portrait}}

\vspace{-0.2cm}
\caption{\footnotesize Visual comparisons on the human $\leftrightarrow$ Portrait task.}\label{fig:Portrait}
\vspace{-3mm} 
\end{figure*}

\begin{figure}[t]
\centering
 \raisebox{0.005\textwidth }{\makebox[0.99\textwidth][l]{ \footnotesize \hspace{ 5pt}  Input \hspace{ 20pt}  Ours \hspace{ 13pt}  w/o drawer \hspace{ 8pt}  w/o $\mathcal{L}_{s}$ \hspace{ 8pt} w/o $\mathcal{L}_{c}$  \hspace{ 2pt}  patch Dis\cite{pix2pix2016} }}
 
 \includegraphics[width=0.16\textwidth,height=0.16\textwidth]{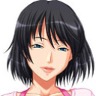}
  \hspace{-5pt}
 \includegraphics[width=0.16\textwidth,height=0.16\textwidth]{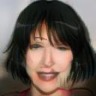}
  \hspace{-6pt}
 \includegraphics[width=0.16\textwidth,height=0.16\textwidth]{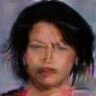}
  \hspace{-6pt}  
 \includegraphics[width=0.16\textwidth,height=0.16\textwidth]{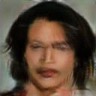}
  \hspace{-6pt}
  \includegraphics[width=0.16\textwidth,height=0.16\textwidth]{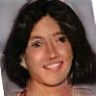}
   \hspace{-6pt}
   \includegraphics[width=0.16\textwidth,height=0.16\textwidth]{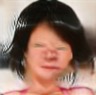}

\vspace{1pt}

 \includegraphics[width=0.16\textwidth,height=0.16\textwidth]{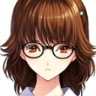}
  \hspace{-5pt}
 \includegraphics[width=0.16\textwidth,height=0.16\textwidth]{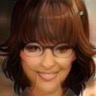}
  \hspace{-6pt}
 \includegraphics[width=0.16\textwidth,height=0.16\textwidth]{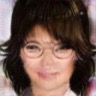}
  \hspace{-6pt}
 \includegraphics[width=0.16\textwidth,height=0.16\textwidth]{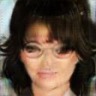}
 \hspace{-6pt}
 \includegraphics[width=0.16\textwidth,height=0.16\textwidth]{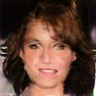}
  \hspace{-6pt} 
 \includegraphics[width=0.16\textwidth,height=0.16\textwidth]{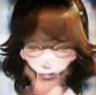}

\vspace{1pt}

 \includegraphics[width=0.16\textwidth,height=0.16\textwidth]{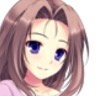}
  \hspace{-5pt}
 \includegraphics[width=0.16\textwidth,height=0.16\textwidth]{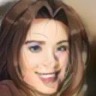}
  \hspace{-6pt}
 \includegraphics[width=0.16\textwidth,height=0.16\textwidth]{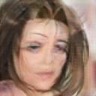}
  \hspace{-6pt}
 \includegraphics[width=0.16\textwidth,height=0.16\textwidth]{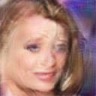} 
 \hspace{-6pt}
 \includegraphics[width=0.16\textwidth,height=0.16\textwidth]{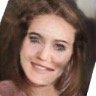}
  \hspace{-6pt} 
 \includegraphics[width=0.16\textwidth,height=0.16\textwidth]{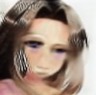}
 
\vspace{1pt}

 \includegraphics[width=0.16\textwidth,height=0.16\textwidth]{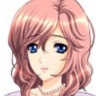}
  \hspace{-5pt}
 \includegraphics[width=0.16\textwidth,height=0.16\textwidth]{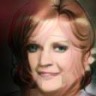}
  \hspace{-6pt}
 \includegraphics[width=0.16\textwidth,height=0.16\textwidth]{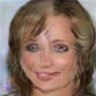}
 \hspace{-6pt}
 \includegraphics[width=0.16\textwidth,height=0.16\textwidth]{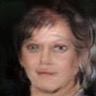}
  \hspace{-6pt}
 \includegraphics[width=0.16\textwidth,height=0.16\textwidth]{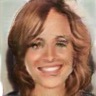}
  \hspace{-6pt}  
  \includegraphics[width=0.16\textwidth,height=0.16\textwidth]{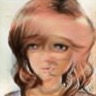}

\vspace{4mm}
\caption{\footnotesize Qualitative ablation study on the Anime $\rightarrow$ human task. The ablation study is performed by removing parts of the network. The patch Dis indicates the model is trained with a patch-GAN \cite{pix2pix2016} instead of our discriminator.}\label{fig:Ablation}
\vspace{-4mm}
\end{figure}

\section{Experiment}

\subsection{Datasets}
We evaluate our model on two types of datasets. The first type of datasets consists of two domains used to demonstrate the effectiveness of the model. These include: 

\bfsection{Anime $\leftrightarrow$ Human} The Anime images are obtained from the Getchu dataset. We use anime-face detector to extract a total of 8,034 anime face images. The human face images are selected from the CelebA dataset. We only use the frontal face images without hats and glasses, and randomly select 14,000 human face images.

\bfsection{Photo $\leftrightarrow$ Portrait}
This dataset is used by DRIT. We follow the train and test split used by the authors. 

The second type of datasets involves multiple image domains to evaluate the efficiency of our model. They include 

\bfsection{Animal Image Translation} To create a multi-domain translation task, we extend the cat$\leftrightarrow$dog dataset used by DRIT with another image collection, the big cat. We collect the big cat images from the ImageNet dataset. 

\bfsection{Face Aging Dataset}
The dataset is provided in~\cite{zhang2017age}. The age is divided into ten categories, i.e., 0-5, 6-10, 11-15, 16-20, 21-30, 31-40, 41-50, 51-60, 61-70, and 71-80. Therefore, we can consider each of the 10 elements as one image domain. In total, there are 10 domains involved.

\bfsection{Collection Style Transfer} 
We use the dataset from CycleGAN \cite{CycleGAN2017}, which contains Monet, Cezanne, Van Gogh, Ukiyo-e, and natural scene images. We train our model once to learn all the mapping among these 5 image domains.


\begin{table}[!b]\setlength{\tabcolsep}{5pt}
\scriptsize 
	\RawFloats
    \caption{\footnotesize Quantitative comparison. We use FID (lower is better) and user study (higher is better) to evaluate the quality of generated images.}
	\label{table1}	
	\begin{center}
	  \begin{tabular}{@{\extracolsep{0.5pt}}l||cccccccc@{}} 
			\hlineB{2}
			\multirow{2}{*}{Setting} & \multicolumn{2}{c}{Human2Anime}&\multicolumn{2}{c}{Anime2Human}&\multicolumn{2}{c}{photo2portrait}&\multicolumn{2}{c}{portrait2photo} \\			\cline{2-3}\cline{4-5}\cline{6-7}\cline{8-9}
			 &FID&AMT&FID&AMT& FID&AMT&FID&AMT\\
			\hline
			real data&1.84&0.50&2.43&0.50&1.12&0.50&2.35&0.50\\
			\hline
			CycleGAN        &17.37&0.17&38.74&0.00&10.13&0.01&18.46&0.09\\
		    DRIT            &23.61&0.01&37.56&0.02&8.23&0.21&20.63&0.03\\
			StarGAN         &22.18&0.01&32.57&0.00&9.26&0.02&19.14&0.08\\
			MUNIT           &17.34&0.10&39.67&0.05&9.45&0.07&20.54&0.12\\
            FUNIT           &22.45&0.01&27.25&0.12&9.84&0.01&24.25&0.02\\
		    Ours            &15.47&0.20&24.47&0.31&8.97&0.18&17.12&0.16\\ \hline   
            w/o drawer      &18.34&-&28.57&-&9.54 & -     &17.81&-\\
            w/o $\mathcal{L}_{c}$ &16.25&-&26.13&-&10.23&-      &17.47&-\\
            w/o $\mathcal{L}_{s}$ &17.34&-&27.63&-&9.85 & -     &17.24&-\\
            patch Dis \cite{pix2pix2016} &21.44&-&31.32&-&11.54 & -     &20.19&-\\            
		    \hlineB{2}
		\end{tabular}
	\end{center}
\vspace{-0.3cm}	
\end{table}

\begin{figure*}[t]
\centering
\includegraphics[width=1.0\linewidth]{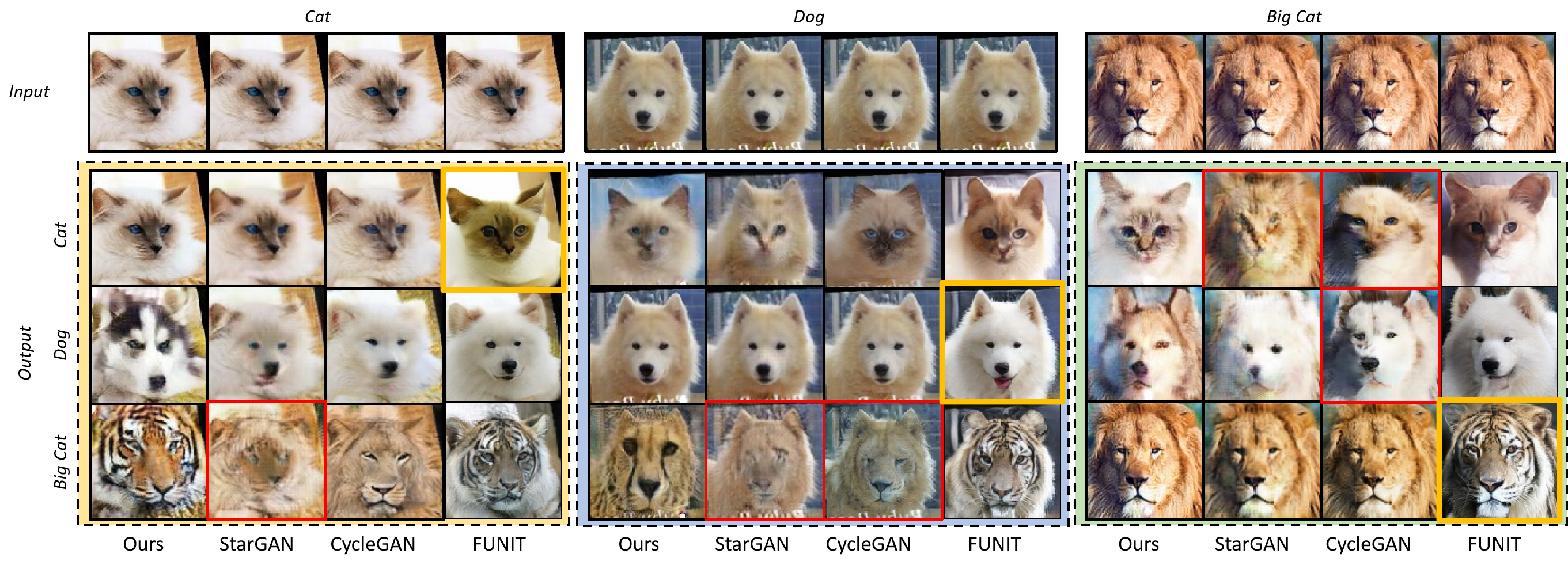}  
\caption{Results on animal dataset. The red rectangles highlight the failure cases in translation, while the yellow rectangles highlight the failure cases in reconstruction.}\label{fig:Animal}
\vspace{-0.3cm} 
\end{figure*}

\subsection{Baseline Methods}
We compare our model with different types of state-of-the-art-models:
CycleGAN~\cite{CycleGAN2017}, DRIT~\cite{lee2018diverse}, MUNIT~\cite{huang2018multimodal}, StarGAN~\cite{choi2018stargan} and FUNIT~\cite{liu2019few}. All of these methods can perform unsupervised image-to-image translation without paired training dataset. In particular, StarGAN and FUNIT belong to multi-domain image translation methods. DRIT and MUNIT are multimodal models. These two models generate diverse samples that are not faithful to the input image. Thus, they are not compared in the multi-domain image translation tasks. We train these baseline models using their public implementation.

\setlength{\tabcolsep}{13pt}
\begin{table}[t!]
\footnotesize 
\RawFloats
\centering
\caption{ Quantitative comparison on the Animal set. The FID and CA scores of the FUNIT are not reported by the authors.}
\label{tab:Animal}
\begin{tabular}{l||cccc}
\hlineB{2}
Setting & AMT & IS & FID& CA \\
\hline
real data &0.500&2.93&0.47&0.99\\
\hline
CycleGAN &0.08&2.26 &25.74&0.82\\
StarGAN& 0.02&1.92 &33.57&0.68\\
FUNIT& 0.24&2.82 &-&-\\
Ours& 0.16 & 2.62&17.47&0.93\\  
\hline
w/o drawer      &-&2.47&19.54      &0.78\\
w/o $\mathcal{L}_{c}$ &-&2.51&18.31      &0.83\\
w/o $\mathcal{L}_{s}$ &-&2.53&18.15      &0.84\\
patch Dis \cite{pix2pix2016} &-&2.36&20.45     &0.69\\  
\hlineB{2}
\end{tabular}
\end{table}

\subsection{Evaluation Metrics}
In unsupervised image-to-image translation, there is no per-pixel ground truth. We utilize the following metrics for evaluation: the Classification Accuracy (CA) and Inception Score (IS) \cite{salimans2016improved}, which is designed to assess multi-classes generation by measuring the probability of the generated images belonging to the distribution of the real data; the FID score \cite{heusel2017gans} and AMT perceptual test \cite{zhang2016colorful} are used to measure the visual quality of the generated images. 

\subsection{Implementation Details}
All the models are trained using Adam optimizer \cite{kingma2014adam} with $\beta_1 = 0.5$ and $\beta_2 = 0.999$, and we train our model for 400,000 iterations with a batch size 16, learning rate 0.0001 and the learning rate is reduced by a factor of 10 after 200,000 iterations. In our experiments, We set $\lambda_{gp} = 10$, $\lambda_1 = 1$, $\lambda_2 =0.5$ and $\mathcal{L}^{identiy}_{1} =10$ and $\mathcal{L}_{vgg} =0.5$ for all of the tasks. For baseline methods, we use their original default parameters for training.

\subsection{Transformation between Two Domains}
We first analyze model effectiveness on the unsupervised tasks involving two domains. These are photo$\leftrightarrow$portrait and human$\leftrightarrow$anime.


\bfsection{Qualitative Comparison}
Learning a mapping between these domains is a more challenging task than facial attribute translation due to the domain difference between these different types of faces. As illustrated in Figure \ref{fig:Anime}, the synthesis of StarGAN and CycleGAN suffers from artifacts and has noticeable trace of the source objects and textures. With the cycle-consistency constraint, the generated images are able to preserve the head direction but not successfully translated to the target domain. We can observe a big improvement in our approach for head pose and color consistency, and promote a large degree of shape deformation such as changing the vertical height of the faces. The multimodal methods, MUNIT and DRIT, yield diverse samples without preserving the color and pose of input images. The comparison on photo$\leftrightarrow$portrait task is shown in Figure \ref{fig:Portrait}, StarGAN, and CycleGAN generate recognizable results in the target domain due to the similarity between these two domains. Our approach achieves competitive results with all of the baselines.

\bfsection{Quantitative Comparison} To quantify the performance of visual quality, we first compute the FID scores between the generated data and the real data. Table \ref{table1} lists the quantitative comparisons. Overall, our model achieves a large degree of shape transformation on a challenging dataset. In particular, even though CycleGAN obtains reasonable performance in translating human to anime, it fails to translate the anime to human, indicating a limitation of dealing with large domain gaps. We proceed to conduct a user study to evaluate the proposed algorithm against the state-of-the-art style transfer methods. Participants are shown the translated images and asked to select the best-translated images. All test images of each dataset are rated by different participants. Our method obtains the best scores.

\begin{figure*}[h!]
\centering
 \raisebox{0.005\textwidth }{\makebox[0.99\textwidth][l]{ \scriptsize \hspace{ 20pt}  Input \hspace{ 30pt}  \hspace{ 26pt}  \hspace{ 31pt}  \hspace{ 76pt}   \hspace{ 24pt}  Ours }}
 \includegraphics[width=0.08\textwidth,height=0.08\textwidth]{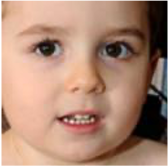}
  \hspace{-3pt}
 \includegraphics[width=0.08\textwidth,height=0.08\textwidth]{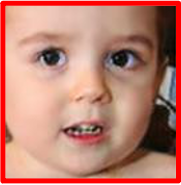}
  \hspace{-5pt}
 \includegraphics[width=0.08\textwidth,height=0.08\textwidth]{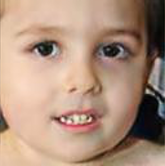}
  \hspace{-5pt}
 \includegraphics[width=0.08\textwidth,height=0.08\textwidth]{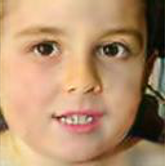}
  \hspace{-5pt}  
 \includegraphics[width=0.08\textwidth,height=0.08\textwidth]{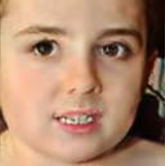}
  \hspace{-5pt}
  \includegraphics[width=0.08\textwidth,height=0.08\textwidth]{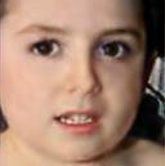}
   \hspace{-5pt}
   \includegraphics[width=0.08\textwidth,height=0.08\textwidth]{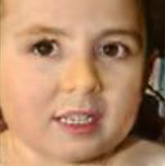}
  \hspace{-5pt}
 \includegraphics[width=0.08\textwidth,height=0.08\textwidth]{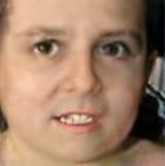}
  \hspace{-5pt}
 \includegraphics[width=0.08\textwidth,height=0.08\textwidth]{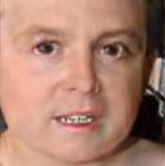}
  \hspace{-5pt}  
 \includegraphics[width=0.08\textwidth,height=0.08\textwidth]{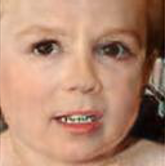}
  \hspace{-5pt}
  \includegraphics[width=0.08\textwidth,height=0.08\textwidth]{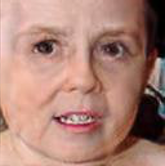}
 \raisebox{0.03\textwidth }{
 \hspace{-8pt}
  \begin{turn}{90} 
 \makebox[0.01\textwidth]{\centering \scriptsize Ours}
\end{turn} }

\vspace{2pt}

 \includegraphics[width=0.08\textwidth,height=0.08\textwidth]{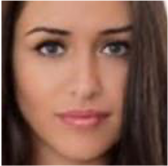}
  \hspace{-3pt}
 \includegraphics[width=0.08\textwidth,height=0.08\textwidth]{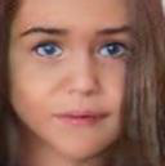}
  \hspace{-5pt}
 \includegraphics[width=0.08\textwidth,height=0.08\textwidth]{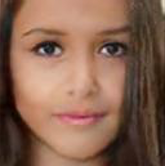}
  \hspace{-5pt}
 \includegraphics[width=0.08\textwidth,height=0.08\textwidth]{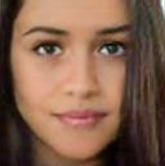}
  \hspace{-5pt}  
 \includegraphics[width=0.08\textwidth,height=0.08\textwidth]{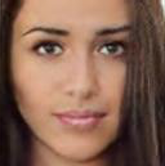}
  \hspace{-5pt}
  \includegraphics[width=0.08\textwidth,height=0.08\textwidth]{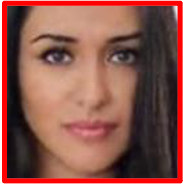}
   \hspace{-5pt}
   \includegraphics[width=0.08\textwidth,height=0.08\textwidth]{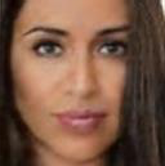}
  \hspace{-5pt}
 \includegraphics[width=0.08\textwidth,height=0.08\textwidth]{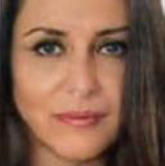}
  \hspace{-5pt}
 \includegraphics[width=0.08\textwidth,height=0.08\textwidth]{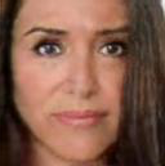}
  \hspace{-5pt}  
 \includegraphics[width=0.08\textwidth,height=0.08\textwidth]{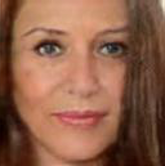}
  \hspace{-5pt}
  \includegraphics[width=0.08\textwidth,height=0.08\textwidth]{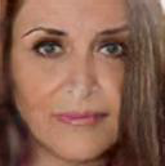}
 \raisebox{0.03\textwidth }{
 \hspace{-8pt}
  \begin{turn}{90} 
 \makebox[0.01\textwidth]{\centering \scriptsize StarGAN}
\end{turn} }

\vspace{2pt}

 \includegraphics[width=0.08\textwidth,height=0.08\textwidth]{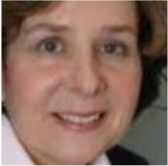}
  \hspace{-3pt}
 \includegraphics[width=0.08\textwidth,height=0.08\textwidth]{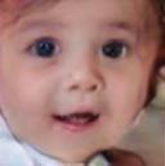}
  \hspace{-5pt}
 \includegraphics[width=0.08\textwidth,height=0.08\textwidth]{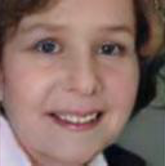}
  \hspace{-5pt}
 \includegraphics[width=0.08\textwidth,height=0.08\textwidth]{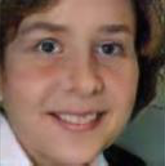}
  \hspace{-5pt}  
 \includegraphics[width=0.08\textwidth,height=0.08\textwidth]{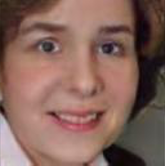}
  \hspace{-5pt}
  \includegraphics[width=0.08\textwidth,height=0.08\textwidth]{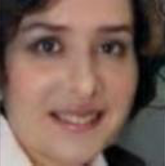}
   \hspace{-5pt}
   \includegraphics[width=0.08\textwidth,height=0.08\textwidth]{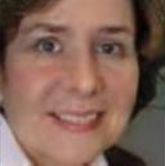}
  \hspace{-5pt}
 \includegraphics[width=0.08\textwidth,height=0.08\textwidth]{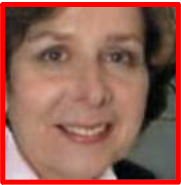}
  \hspace{-5pt}
 \includegraphics[width=0.08\textwidth,height=0.08\textwidth]{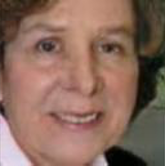}
  \hspace{-5pt}  
 \includegraphics[width=0.08\textwidth,height=0.08\textwidth]{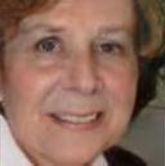}
  \hspace{-5pt}
  \includegraphics[width=0.08\textwidth,height=0.08\textwidth]{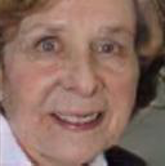}
 \raisebox{0.03\textwidth }{
 \hspace{-8pt}
  \begin{turn}{90} 
 \makebox[0.01\textwidth]{\centering \scriptsize CAAE}
\end{turn} }

\vspace{2pt}

 \includegraphics[width=0.08\textwidth,height=0.08\textwidth]{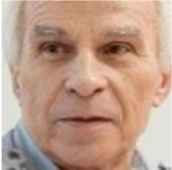}
  \hspace{-3pt}
 \includegraphics[width=0.08\textwidth,height=0.08\textwidth]{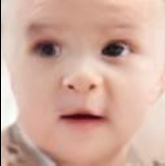}
  \hspace{-5pt}
 \includegraphics[width=0.08\textwidth,height=0.08\textwidth]{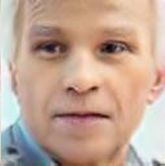}
  \hspace{-5pt}
 \includegraphics[width=0.08\textwidth,height=0.08\textwidth]{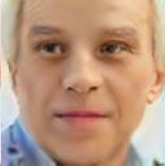}
  \hspace{-5pt}  
 \includegraphics[width=0.08\textwidth,height=0.08\textwidth]{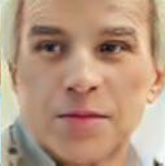}
  \hspace{-5pt}
  \includegraphics[width=0.08\textwidth,height=0.08\textwidth]{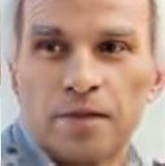}
   \hspace{-5pt}
   \includegraphics[width=0.08\textwidth,height=0.08\textwidth]{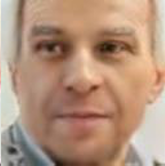}
  \hspace{-5pt}
 \includegraphics[width=0.08\textwidth,height=0.08\textwidth]{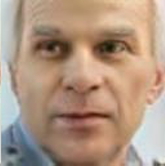}
  \hspace{-5pt}
 \includegraphics[width=0.08\textwidth,height=0.08\textwidth]{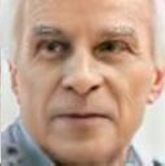}
  \hspace{-5pt}  
 \includegraphics[width=0.08\textwidth,height=0.08\textwidth]{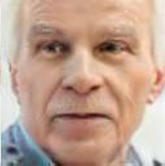}
  \hspace{-5pt}
  \includegraphics[width=0.08\textwidth,height=0.08\textwidth]{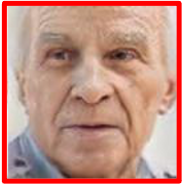}
 \raisebox{0.03\textwidth }{
 \hspace{-8pt}
  \begin{turn}{90} 
 \makebox[0.01\textwidth]{\centering \scriptsize CAAE}
\end{turn} }

 \raisebox{0.005\textwidth }{\makebox[0.99\textwidth][l]{ \scriptsize \hspace{ 25pt}   \hspace{ 38pt}  0-5 \hspace{ 26pt}  6-10 \hspace{ 26pt}  11-15\hspace{ 26pt} 16-20  \hspace{ 24pt}  21-30 \hspace{ 24pt}31-40\hspace{ 24pt}41-50\hspace{ 26pt}51-60\hspace{ 26pt}61-70\hspace{ 26pt}71-80}}
\caption{Samples generated by our method. The  red  rectangles  indicate  the  input images are belong to the corresponding age range.}
\label{vis:face}
\vspace{-3mm}
\end{figure*}

\begin{figure*}[h!]
\centering
 \raisebox{0.005\textwidth }{\makebox[0.99\textwidth][l]{ \scriptsize \hspace{ 20pt}  Input }}
 \includegraphics[width=0.08\textwidth,height=0.08\textwidth]{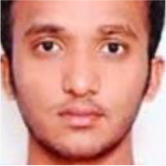}
  \hspace{-3pt}
 \includegraphics[width=0.08\textwidth,height=0.08\textwidth]{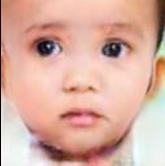}
  \hspace{-5pt}
 \includegraphics[width=0.08\textwidth,height=0.08\textwidth]{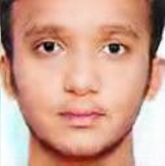}
  \hspace{-5pt}
 \includegraphics[width=0.08\textwidth,height=0.08\textwidth]{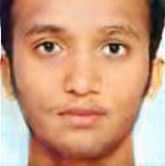}
  \hspace{-5pt}  
 \includegraphics[width=0.08\textwidth,height=0.08\textwidth]{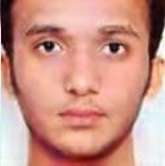}
  \hspace{-5pt}
  \includegraphics[width=0.08\textwidth,height=0.08\textwidth]{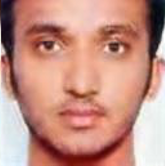}
   \hspace{-5pt}
   \includegraphics[width=0.08\textwidth,height=0.08\textwidth]{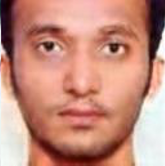}
  \hspace{-5pt}
 \includegraphics[width=0.08\textwidth,height=0.08\textwidth]{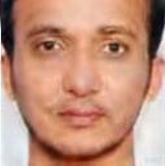}
  \hspace{-5pt}
 \includegraphics[width=0.08\textwidth,height=0.08\textwidth]{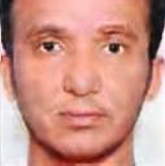}
  \hspace{-5pt}  
 \includegraphics[width=0.08\textwidth,height=0.08\textwidth]{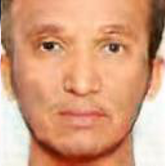}
  \hspace{-5pt}
  \includegraphics[width=0.08\textwidth,height=0.08\textwidth]{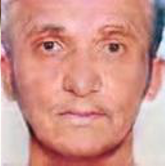}
 \raisebox{0.03\textwidth }{
 \hspace{-8pt}
  \begin{turn}{90} 
 \makebox[0.01\textwidth]{\centering \scriptsize Ours}
\end{turn} }

\vspace{2pt}

 \includegraphics[width=0.08\textwidth,height=0.08\textwidth]{fig-age/c1}
  \hspace{-3pt}
 \includegraphics[width=0.08\textwidth,height=0.08\textwidth]{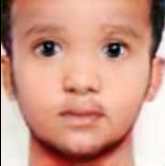}
  \hspace{-5pt}
 \includegraphics[width=0.08\textwidth,height=0.08\textwidth]{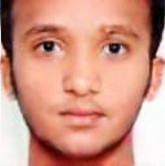}
  \hspace{-5pt}
 \includegraphics[width=0.08\textwidth,height=0.08\textwidth]{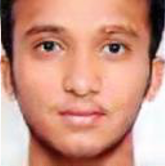}
  \hspace{-5pt}  
 \includegraphics[width=0.08\textwidth,height=0.08\textwidth]{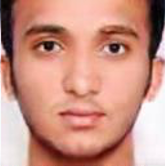}
  \hspace{-5pt}
  \includegraphics[width=0.08\textwidth,height=0.08\textwidth]{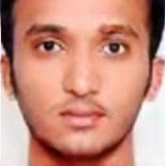}
   \hspace{-5pt}
   \includegraphics[width=0.08\textwidth,height=0.08\textwidth]{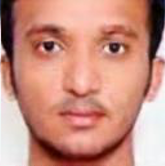}
  \hspace{-5pt}
 \includegraphics[width=0.08\textwidth,height=0.08\textwidth]{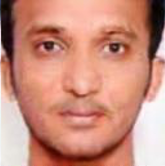}
  \hspace{-5pt}
 \includegraphics[width=0.08\textwidth,height=0.08\textwidth]{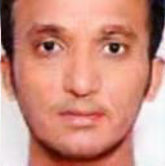}
  \hspace{-5pt}  
 \includegraphics[width=0.08\textwidth,height=0.08\textwidth]{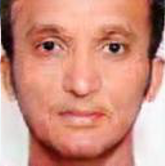}
  \hspace{-5pt}
  \includegraphics[width=0.08\textwidth,height=0.08\textwidth]{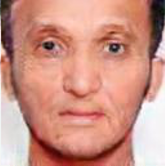}
 \raisebox{0.03\textwidth }{
 \hspace{-8pt}
  \begin{turn}{90} 
 \makebox[0.01\textwidth]{\centering \scriptsize StarGAN}
\end{turn} }

\vspace{2pt}

 \includegraphics[width=0.08\textwidth,height=0.08\textwidth]{fig-age/c1}
  \hspace{-3pt}
 \includegraphics[width=0.08\textwidth,height=0.08\textwidth]{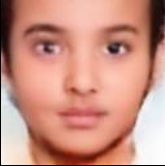}
  \hspace{-5pt}
 \includegraphics[width=0.08\textwidth,height=0.08\textwidth]{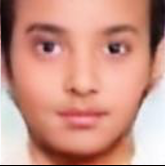}
  \hspace{-5pt}
 \includegraphics[width=0.08\textwidth,height=0.08\textwidth]{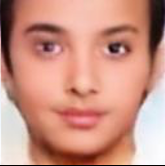}
  \hspace{-5pt}  
 \includegraphics[width=0.08\textwidth,height=0.08\textwidth]{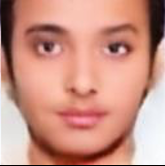}
  \hspace{-5pt}
  \includegraphics[width=0.08\textwidth,height=0.08\textwidth]{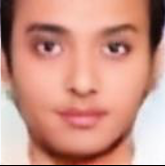}
   \hspace{-5pt}
   \includegraphics[width=0.08\textwidth,height=0.08\textwidth]{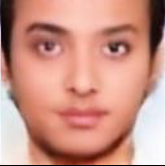}
  \hspace{-5pt}
 \includegraphics[width=0.08\textwidth,height=0.08\textwidth]{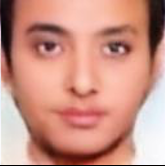}
  \hspace{-5pt}
 \includegraphics[width=0.08\textwidth,height=0.08\textwidth]{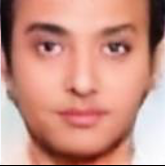}
  \hspace{-5pt}  
 \includegraphics[width=0.08\textwidth,height=0.08\textwidth]{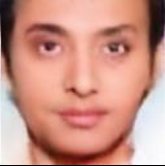}
  \hspace{-5pt}
  \includegraphics[width=0.08\textwidth,height=0.08\textwidth]{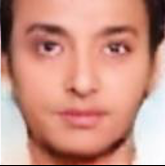}
 \raisebox{0.03\textwidth }{
 \hspace{-8pt}
  \begin{turn}{90} 
 \makebox[0.01\textwidth]{\centering \scriptsize CAAE}
\end{turn} }

 \raisebox{0.005\textwidth }{\makebox[0.99\textwidth][l]{ \scriptsize \hspace{ 25pt}  19 \hspace{ 30pt}  0-5 \hspace{ 26pt}  6-10 \hspace{ 26pt}  11-15\hspace{ 26pt} 16-20  \hspace{ 24pt}  21-30 \hspace{ 24pt}31-40\hspace{ 24pt}41-50\hspace{ 26pt}51-60\hspace{ 26pt}61-70\hspace{ 26pt}71-80}}
\vspace{-0.3cm}   
\caption{\footnotesize Results on face aging dataset.}\label{fig:face}
\end{figure*}

\bfsection{Ablation Study} We perform the ablation study by removing parts of the network to validate the effectiveness of the proposed network. We show the results without the input-output drawer, content perceptual loss $\mathcal{L}_{c}$, or style perceptual loss $\mathcal{L}_{s}$, respectively. Without using the input-output drawer, the performance is significantly degraded as shown in Figure \ref{fig:Ablation}. The content perceptual loss works to preserve visual characteristics of the input image including color and pose, etc. The style perceptual loss increases the domain consistency between the generated images and images in the target domain. Training with the patch discriminator \cite{pix2pix2016} neither exhibits the shape deformation in the content nor faithful style variation. Quantitative results of the FID score is reported in Table \ref{table1}, which indicates that these techniques are important for the generation quality and pose preservation. 


\setlength{\tabcolsep}{4pt}
\begin{table}[t!]
\footnotesize 
\RawFloats
\caption{Quantitative comparison on face aging dataset.}\label{tab:face}
\centering
\begin{tabular}{l||cccccc}
\hlineB{2}
Setting & FID& CA &IS&AMT&Inter-class&Intra-class\\
\hline
CAAE &10.95&0.42 &3.24&0.06&18.44&8.45 \\
StarGAN& 4.18&0.78&4.56&0.13&23.28&5.25\\
Ours& 4.26 & 0.86&5.12&0.31&25.87&4.43\\   
\hline
w/o drawer      &4.55&0.76&4.54 &-&-&-\\
w/o $\mathcal{L}_{c}$ &4.45&0.81&4.63      &-&-&-\\
w/o $\mathcal{L}_{s}$ &4.34&0.79&4.85      &-&-&-\\
patch Dis \cite{pix2pix2016} &5.23&0.69&4.11 &-&-&-\\
\hlineB{2}
\end{tabular}
\end{table}

\subsection{Translations across Multiple Domains}
We proceed to evaluate our model on applications that contain multiple domains. 

\begin{figure}[t]
\begin{subfigure}[b]{0.32\textwidth}
\includegraphics[width=1.\textwidth]{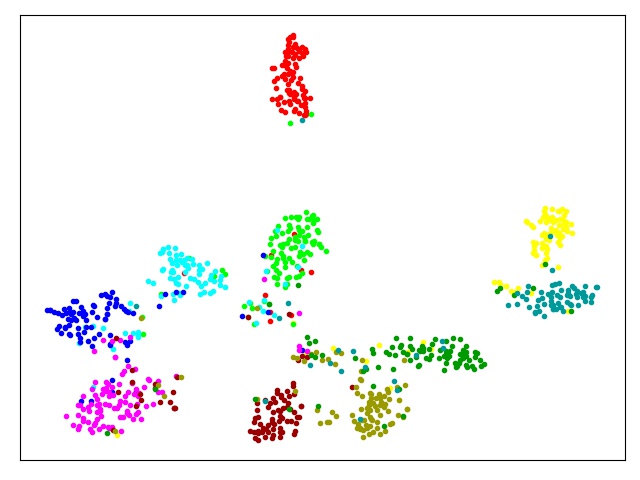}
\caption{Ours}
\end{subfigure}
\begin{subfigure}[b]{0.32\textwidth}
\includegraphics[width=1.\textwidth]{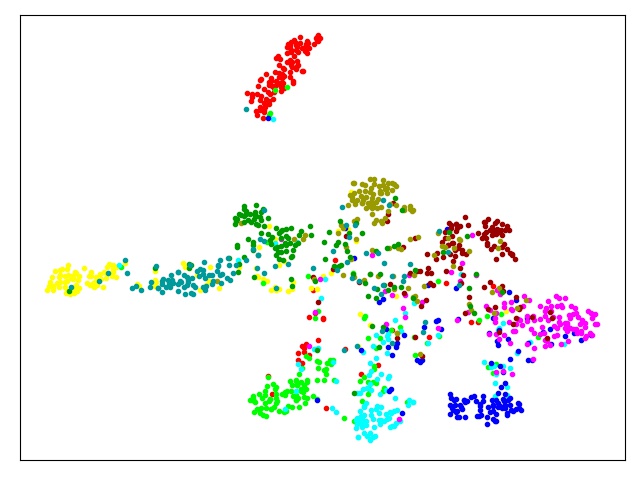}
\caption{StarGAN}
\end{subfigure}
\begin{subfigure}[b]{0.32\textwidth}
\includegraphics[width=1.\textwidth]{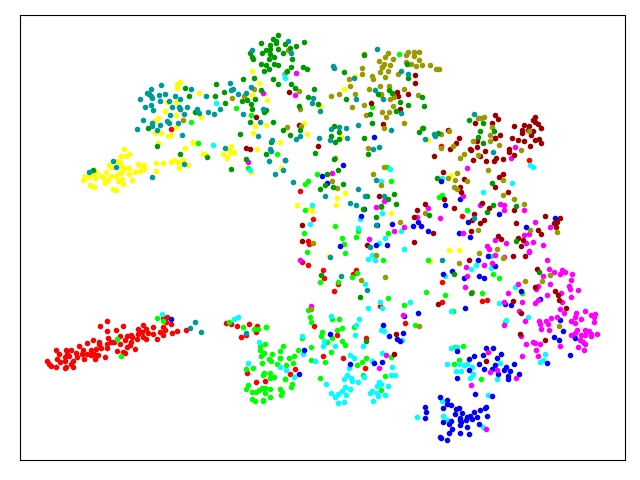}
\caption{CAAE}
\end{subfigure}
\caption{
Feature distribution comparison. The feature is extracted from the pre-trained classification network on the translated images. one color per class.}
\label{fig:tsne}
\end{figure}

\bfsection{Animal Image Translation}
The motivation of our work is to learn a unified model for multi-domains image translation with large domain gaps. We hope our model is both efficient and effective to handle multi-domain image translation tasks. This animal image translation task consisting of three domains is characterized by significant cross-domain shape variation. In Figure \ref{fig:Animal}, we demonstrate the multi-domain translation results of different methods. Note that we do not train the FUNIT model on this dataset, instead, we employ the pre-trained model, which was trained on 139 categories of animal images by the authors. The FUNIT generates appealing results in translation, but the reconstructed animal faces are much different from the original inputs. Our proposed model achieves competitive results against the baselines. First, our model captures domain-related details. Given an input image, the translation outputs cover multiple fine-grained animal characteristics in the target domain. The shape of an animal has undergone significant transformations, but the pose is largely preserved. Second, our model faithfully reconstructs the input images while FUNIT generates a new similar sample.

The quantitative results are listed in Table \ref{tab:Animal}. We compare the overall model capacity with other baselines. The CycleGAN model is inefficient for tasks involving multiple domains. For this task, it needs to train $3$ times separately to cover all the mappings between each domain pair. The StarGAN is efficient in learning a unified mapping. However, it fails to disentangle domain-specific features. Our model advances in translating domain-consistent textures across multiple domains. Overall, the FUNIT achieves the best metric scores, we consider this as the benefits of large numbers of training data across the 139 domains. 


\begin{figure}[t!]
\centering
 \includegraphics[width=0.16\textwidth,height=0.16\textwidth]{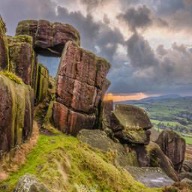}
  \hspace{-5pt}
 \includegraphics[width=0.16\textwidth,height=0.16\textwidth]{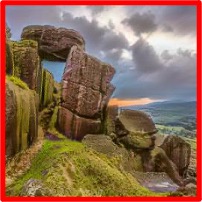}
  \hspace{-6pt}
 \includegraphics[width=0.16\textwidth,height=0.16\textwidth]{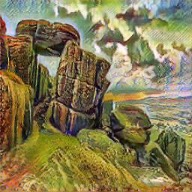}
  \hspace{-6pt}  
 \includegraphics[width=0.16\textwidth,height=0.16\textwidth]{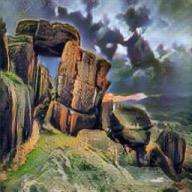}
  \hspace{-6pt}
  \includegraphics[width=0.16\textwidth,height=0.16\textwidth]{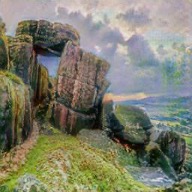}
   \hspace{-6pt}
   \includegraphics[width=0.16\textwidth,height=0.16\textwidth]{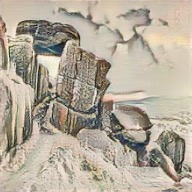}

\vspace{1pt}

 \includegraphics[width=0.16\textwidth,height=0.16\textwidth]{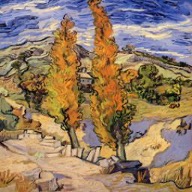}
  \hspace{-5pt}
 \includegraphics[width=0.16\textwidth,height=0.16\textwidth]{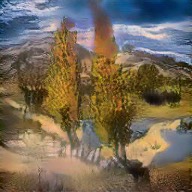}
  \hspace{-6pt}
 \includegraphics[width=0.16\textwidth,height=0.16\textwidth]{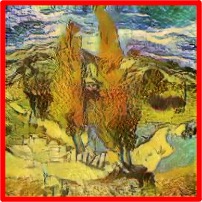}
  \hspace{-6pt}
 \includegraphics[width=0.16\textwidth,height=0.16\textwidth]{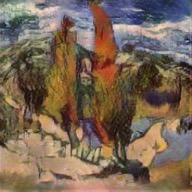}
 \hspace{-6pt}
 \includegraphics[width=0.16\textwidth,height=0.16\textwidth]{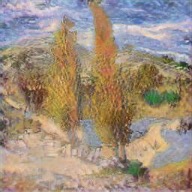}
  \hspace{-6pt} 
 \includegraphics[width=0.16\textwidth,height=0.16\textwidth]{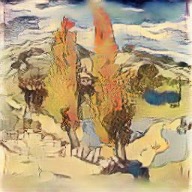}

\vspace{1pt}

 \includegraphics[width=0.16\textwidth,height=0.16\textwidth]{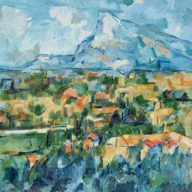}
  \hspace{-5pt}
 \includegraphics[width=0.16\textwidth,height=0.16\textwidth]{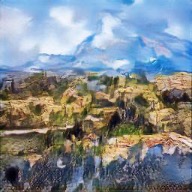}
  \hspace{-6pt}
 \includegraphics[width=0.16\textwidth,height=0.16\textwidth]{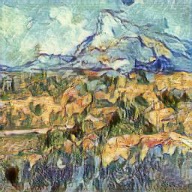}
  \hspace{-6pt}
 \includegraphics[width=0.16\textwidth,height=0.16\textwidth]{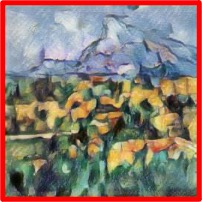} 
 \hspace{-6pt}
 \includegraphics[width=0.16\textwidth,height=0.16\textwidth]{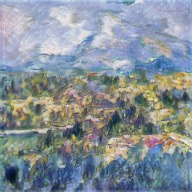}
  \hspace{-6pt} 
 \includegraphics[width=0.16\textwidth,height=0.16\textwidth]{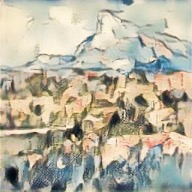}

\vspace{1pt}

 \includegraphics[width=0.16\textwidth,height=0.16\textwidth]{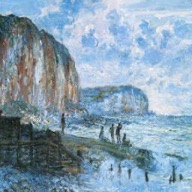}
  \hspace{-5pt}
 \includegraphics[width=0.16\textwidth,height=0.16\textwidth]{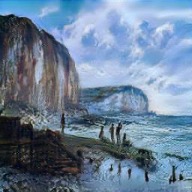}
  \hspace{-6pt}
 \includegraphics[width=0.16\textwidth,height=0.16\textwidth]{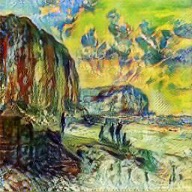}
 \hspace{-6pt}
 \includegraphics[width=0.16\textwidth,height=0.16\textwidth]{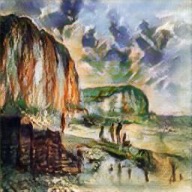}
  \hspace{-6pt}
 \includegraphics[width=0.16\textwidth,height=0.16\textwidth]{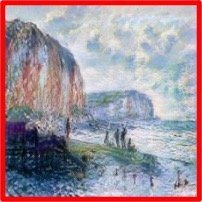}
  \hspace{-6pt}  
  \includegraphics[width=0.16\textwidth,height=0.16\textwidth]{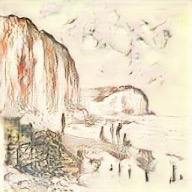}

\vspace{1pt}

 \includegraphics[width=0.16\textwidth,height=0.16\textwidth]{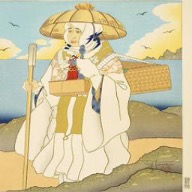}
  \hspace{-5pt}
 \includegraphics[width=0.16\textwidth,height=0.16\textwidth]{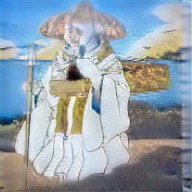}
  \hspace{-6pt}
 \includegraphics[width=0.16\textwidth,height=0.16\textwidth]{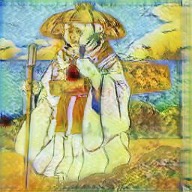}
 \hspace{-6pt}
 \includegraphics[width=0.16\textwidth,height=0.16\textwidth]{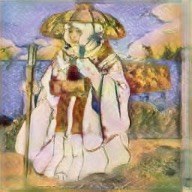}
  \hspace{-6pt}
 \includegraphics[width=0.16\textwidth,height=0.16\textwidth]{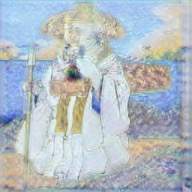}
  \hspace{-6pt}  
  \includegraphics[width=0.16\textwidth,height=0.16\textwidth]{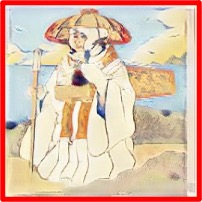}


 \raisebox{0.005\textwidth }{\makebox[0.99\textwidth][l]{ \scriptsize \hspace{ 5pt}  Input \hspace{ 20pt}  Photo \hspace{ 16pt}  Van Gogh \hspace{ 16pt}  Cezanne\hspace{ 16pt} Monet  \hspace{ 16pt}  Ukiyo-e }}
\caption{
The multi-domain translation task of multi-style image generation. The first column is the real images and the rest columns are the generated results in different artistic styles.}\label{fig:painting}
\vspace{-3mm}
\end{figure}

\bfsection{Human Face Aging} The Age Progression/Regression requires the generative model to directly produce images with the desired age attribute. To successfully generate the aging images, this task requires both textures and global shapes transformation. The conditional adversarial autoencoder (CAAE \cite{zhang2017age}) learns a face manifold for age progression and regression. In this work, we consider it as a domain translation task. Instead of learning a latent manifold, we consider each age period as one domain, and translate the input image to the corresponding age domain. We take the StarGAN and the CAAE model as baselines. We present the qualitative performance in Figure \ref{fig:face}. The results show that our model can translate both the texture and shape. It is noteworthy that our mode translates the input image to an infant of age 0-5 with face shape transformation while other baselines suffer the difficulties. We compare the Feature distribution comparison using t-SNE \cite{maaten2008visualizing} in Figure \ref{fig:tsne}. We observe that the translated images generated by our method are clustered in the feature space.

\begin{figure}[t]
\centering
\includegraphics[width=0.7\linewidth]{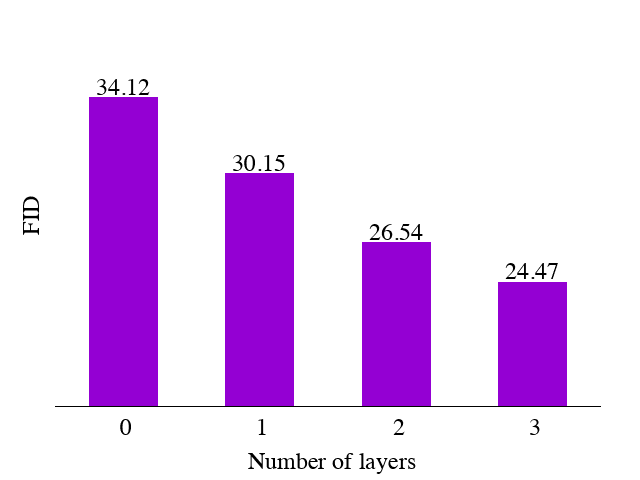}  
\caption{The accuracy (FID score) for the discriminator with respect to different numbers of dilated convolutional layers on the anime to human task.}\label{fig:dilate}
\vspace{-0.3cm} 
\end{figure}

\begin{figure}[t]
\centering
\includegraphics[width=0.8\linewidth]{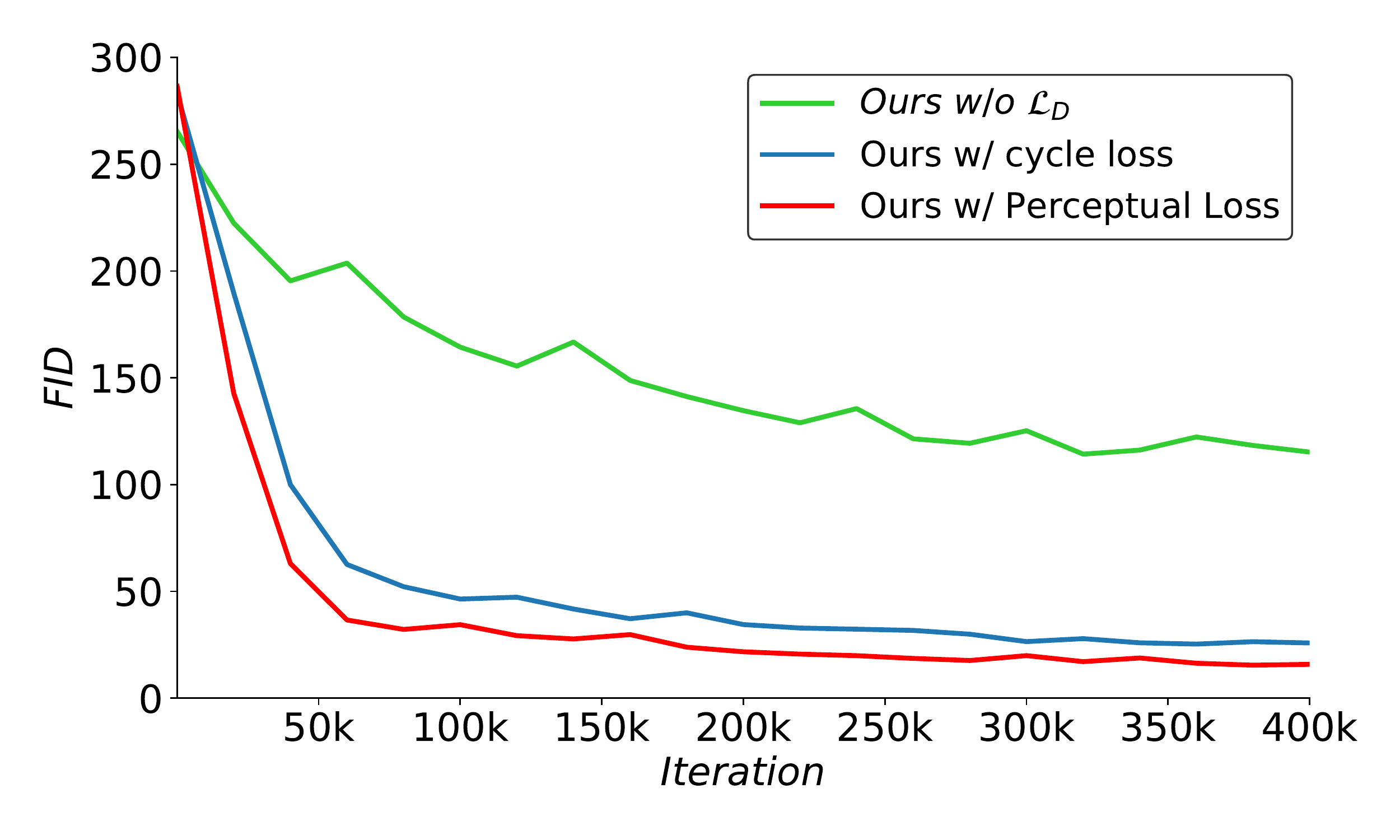}  
\caption{ Quantitative results of the FID scores at the training stage.}\label{fig:curv}
\vspace{-0.3cm} 
\end{figure}

\begin{figure*}[h!]
\centering
 \raisebox{0.005\textwidth }{\makebox[0.114\textwidth]{\centering \scriptsize Input}}
  \hspace{-5pt}   
 \includegraphics[width=0.114\textwidth,height=0.114\textwidth]{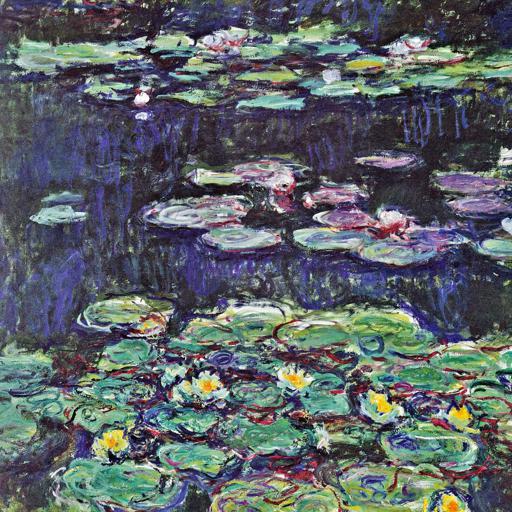}
  \hspace{-5pt}
 \includegraphics[width=0.114\textwidth,height=0.114\textwidth]{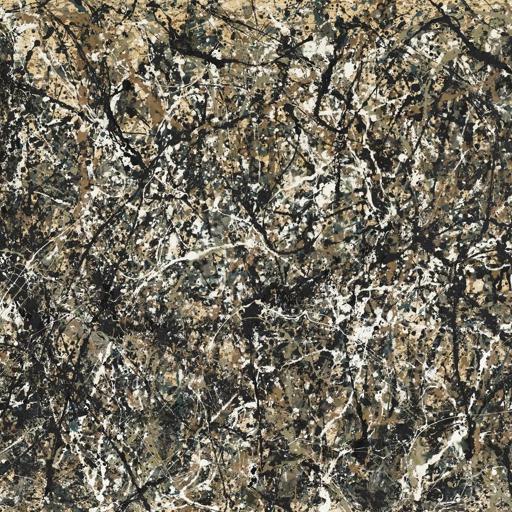}
  \hspace{-5pt}
 \includegraphics[width=0.114\textwidth,height=0.114\textwidth]{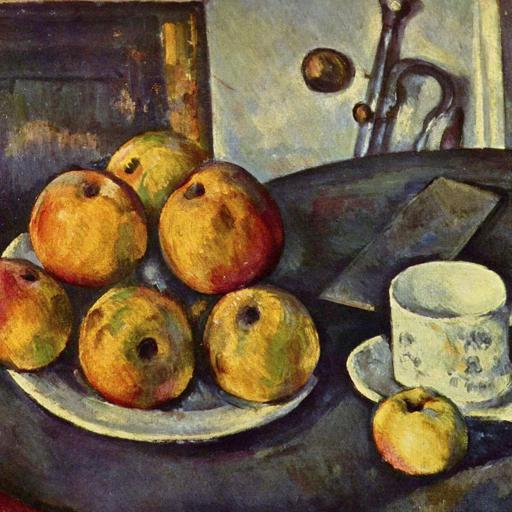}
 \hspace{-4pt}
 \raisebox{0.005\textwidth }{\makebox[0.114\textwidth]{\centering \scriptsize Input}}
  \hspace{-5pt}  
 \includegraphics[width=0.114\textwidth,height=0.114\textwidth]{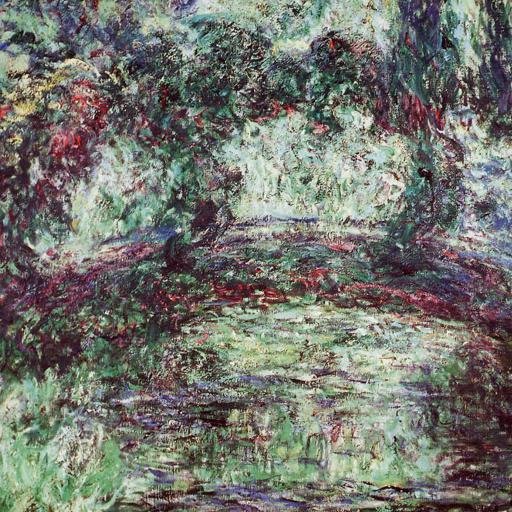}
  \hspace{-5pt}
  \includegraphics[width=0.114\textwidth,height=0.114\textwidth]{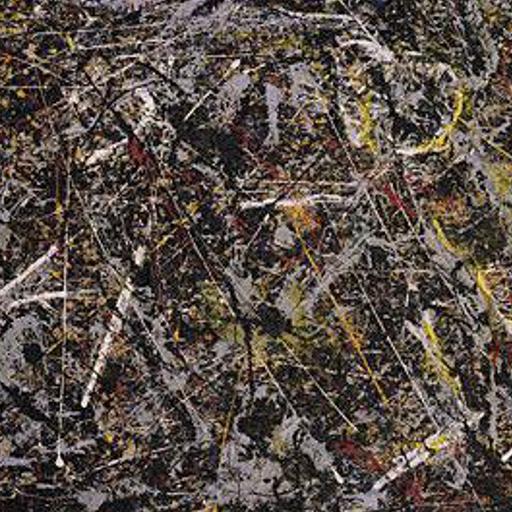}
   \hspace{-5pt}
   \includegraphics[width=0.114\textwidth,height=0.114\textwidth]{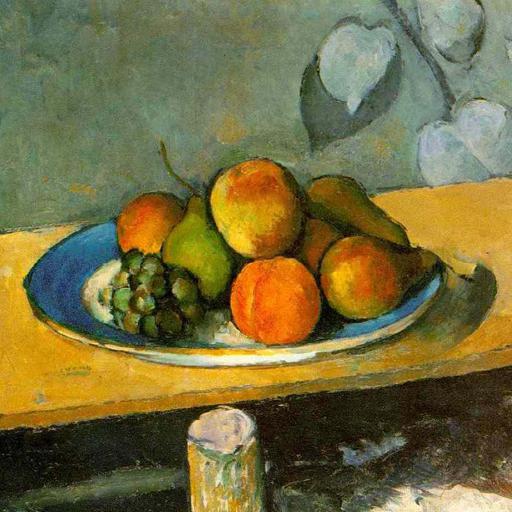}
  \begin{turn}{70} 
 \raisebox{0.04\textwidth }{\makebox[0.05\textwidth]{\centering \scriptsize Style}}
\end{turn} 

\vspace{2pt}

 \includegraphics[width=0.114\textwidth,height=0.114\textwidth]{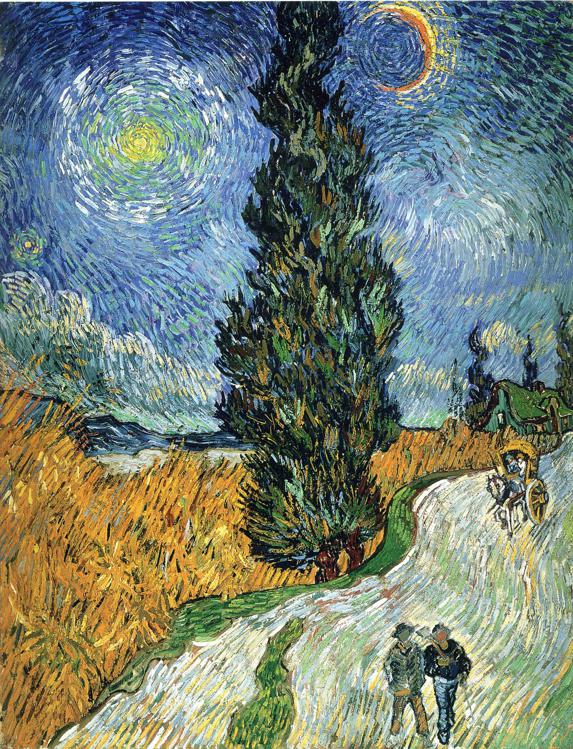}
  \hspace{-5pt}
 \includegraphics[width=0.114\textwidth,height=0.114\textwidth]{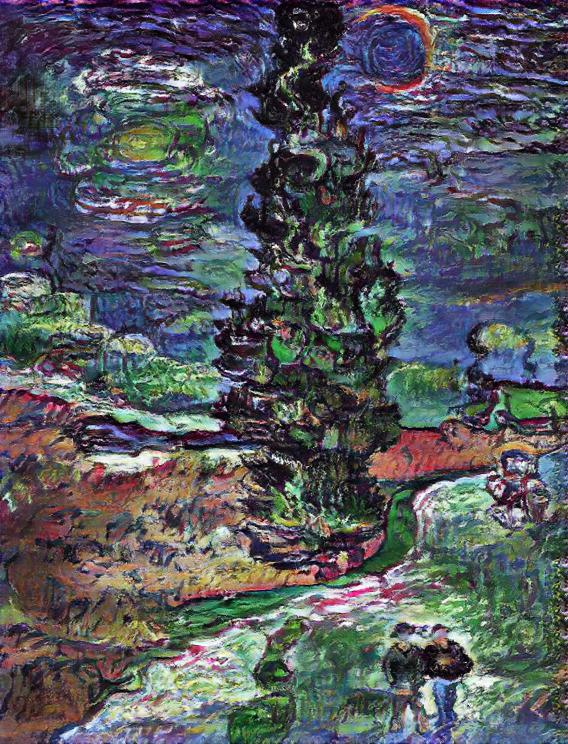}
  \hspace{-5pt}
 \includegraphics[width=0.114\textwidth,height=0.114\textwidth]{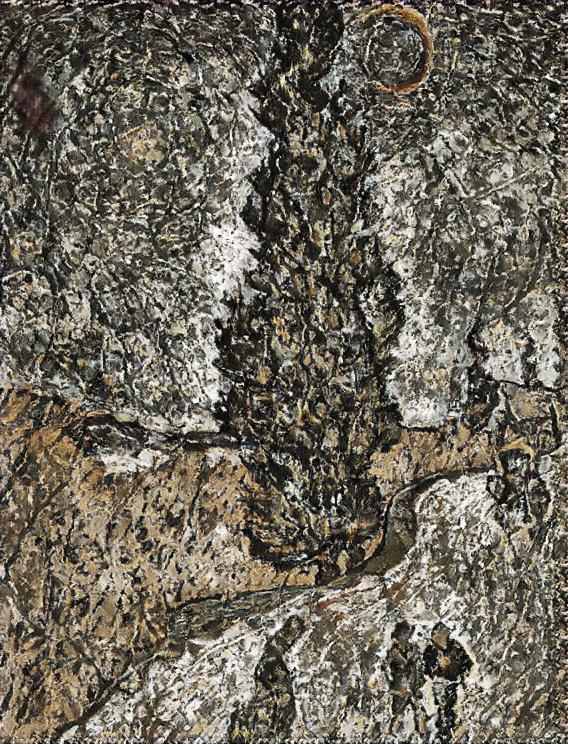}
  \hspace{-5pt}
 \includegraphics[width=0.114\textwidth,height=0.114\textwidth]{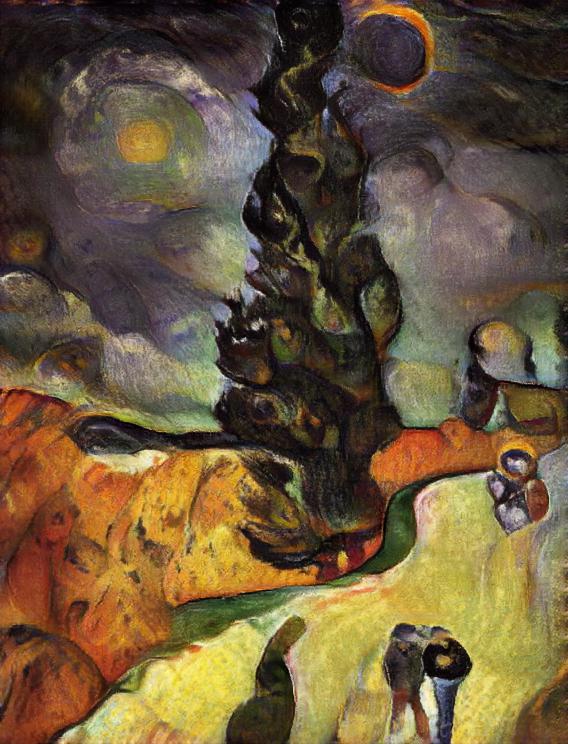}
 \hspace{-4pt}
 \includegraphics[width=0.114\textwidth,height=0.114\textwidth]{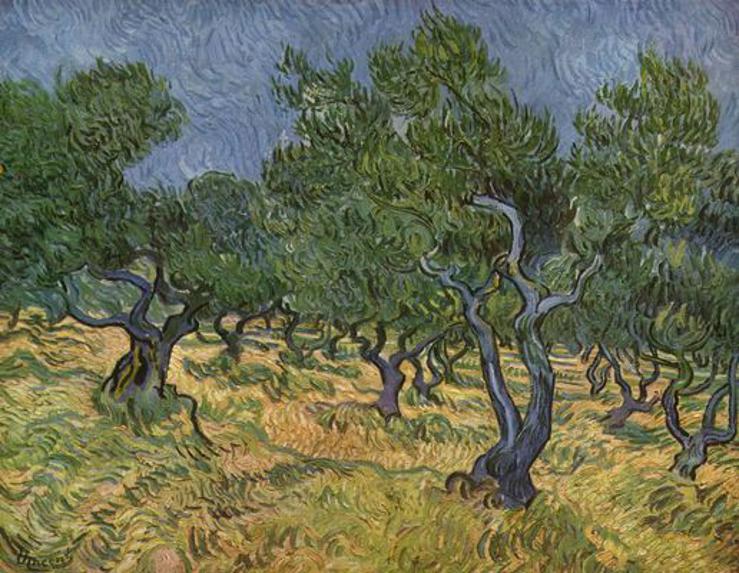}
  \hspace{-5pt} 
 \includegraphics[width=0.114\textwidth,height=0.114\textwidth]{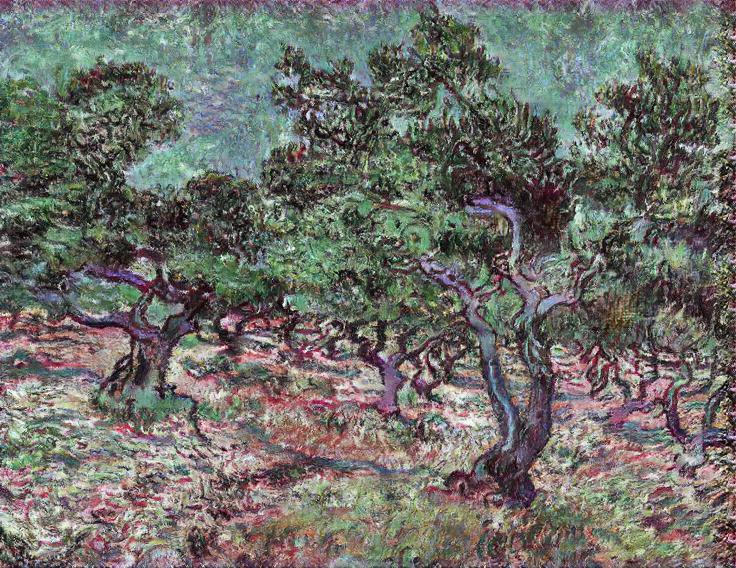}
  \hspace{-5pt}
  \includegraphics[width=0.114\textwidth,height=0.114\textwidth]{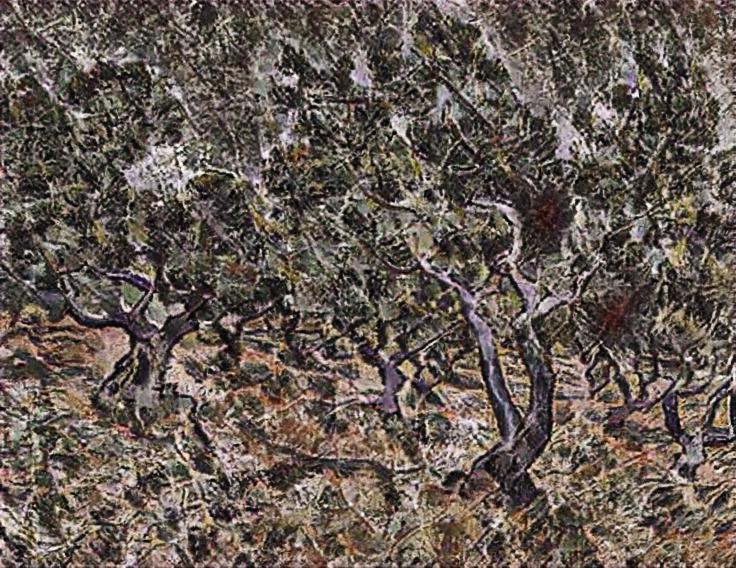}
  \hspace{-5pt}
  \includegraphics[width=0.114\textwidth,height=0.114\textwidth]{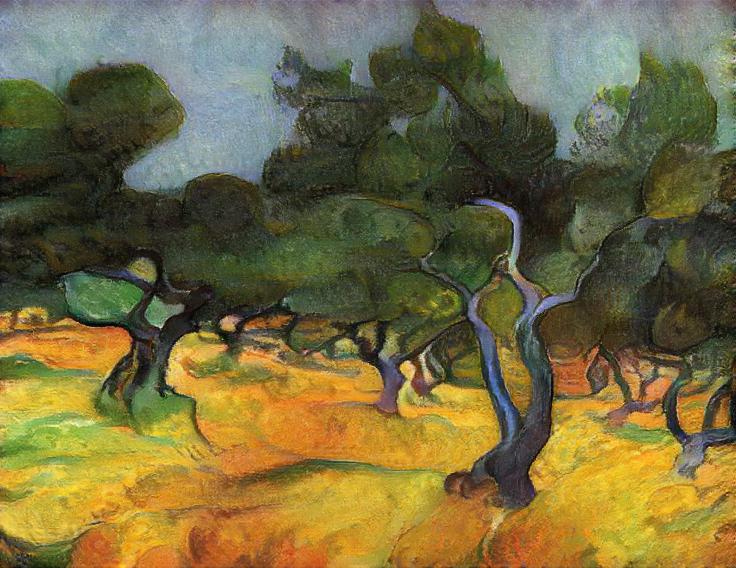}
  \begin{turn}{70} 
 \raisebox{0.04\textwidth }{\makebox[0.05\textwidth]{\centering \scriptsize Ours w/ drawer}}
\end{turn}

\vspace{2pt}

 \includegraphics[width=0.114\textwidth,height=0.114\textwidth]{fig-sty-drawer/c1}
  \hspace{-5pt}
 \includegraphics[width=0.114\textwidth,height=0.114\textwidth]{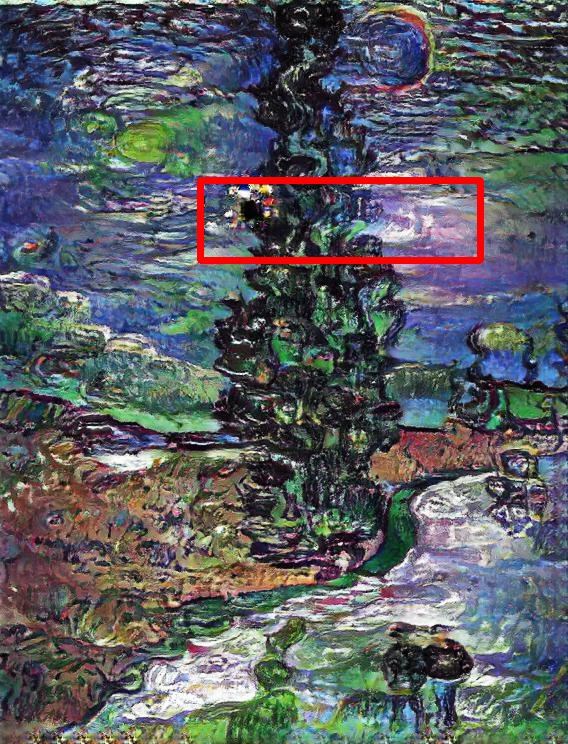}
  \hspace{-5pt}
 \includegraphics[width=0.114\textwidth,height=0.114\textwidth]{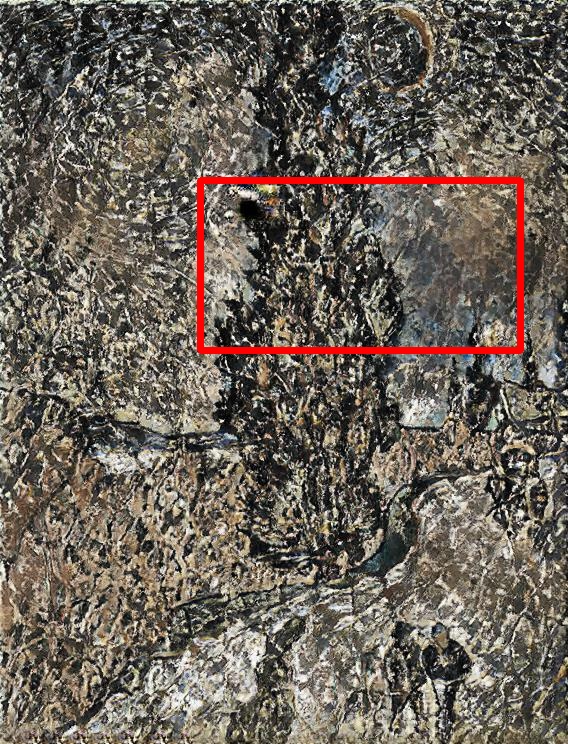}
  \hspace{-5pt}
 \includegraphics[width=0.114\textwidth,height=0.114\textwidth]{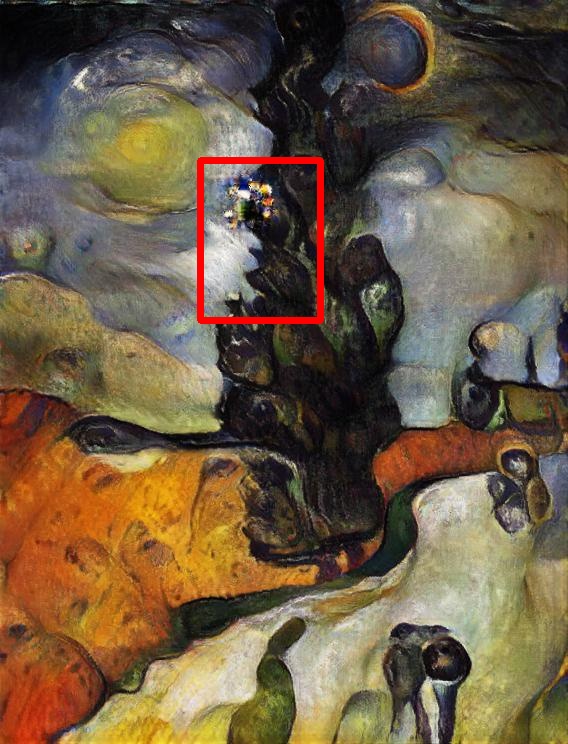} 
 \hspace{-4pt}
 \includegraphics[width=0.114\textwidth,height=0.114\textwidth]{fig-sty-drawer/c2}
  \hspace{-5pt} 
 \includegraphics[width=0.114\textwidth,height=0.114\textwidth]{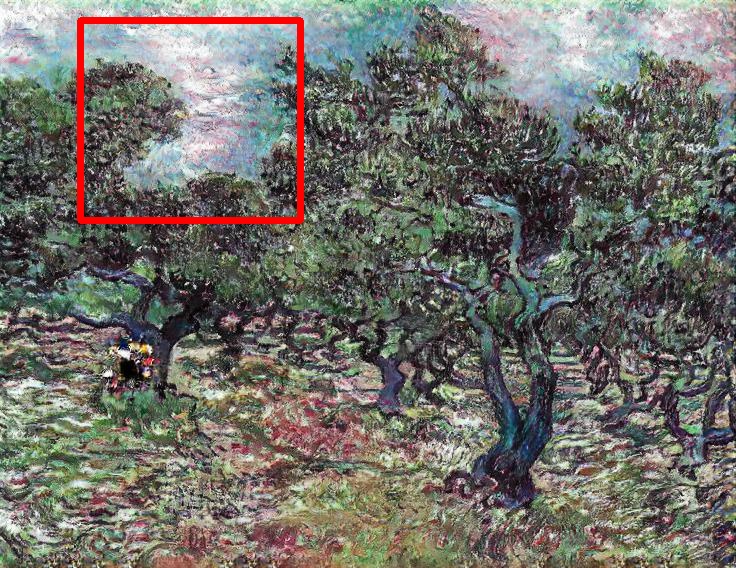}
  \hspace{-5pt}
  \includegraphics[width=0.114\textwidth,height=0.114\textwidth]{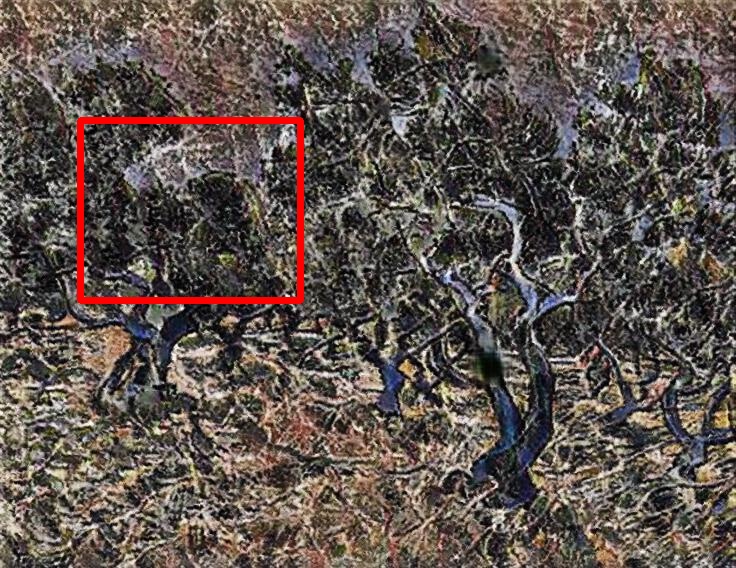}
  \hspace{-5pt}
  \includegraphics[width=0.114\textwidth,height=0.114\textwidth]{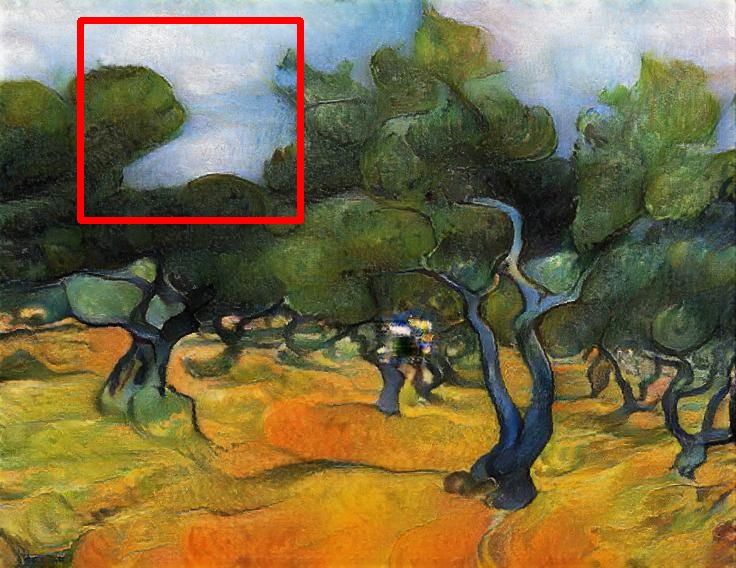}
  \begin{turn}{70} 
 \raisebox{0.04\textwidth }{\makebox[0.05\textwidth]{\centering \scriptsize Ours w/o drawer}}
\end{turn}

\vspace{2pt}

 \includegraphics[width=0.114\textwidth,height=0.114\textwidth]{fig-sty-drawer/c1}
  \hspace{-5pt}
 \includegraphics[width=0.114\textwidth,height=0.114\textwidth]{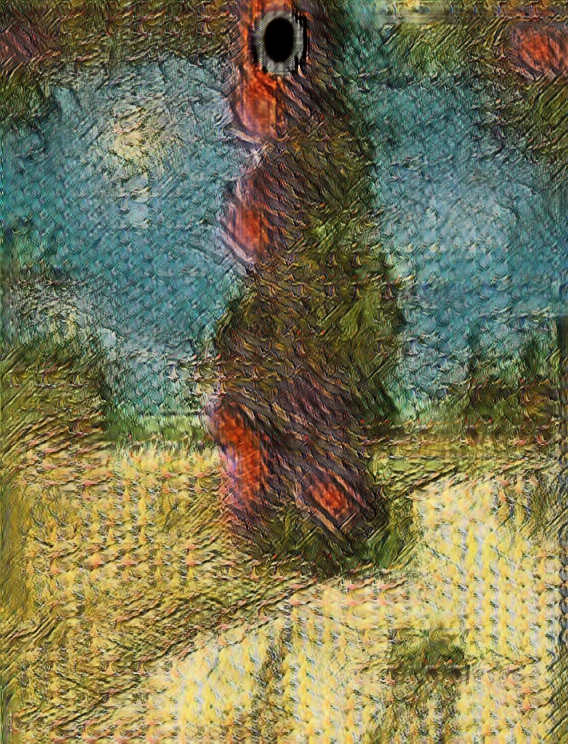}
  \hspace{-5pt}
 \includegraphics[width=0.114\textwidth,height=0.114\textwidth]{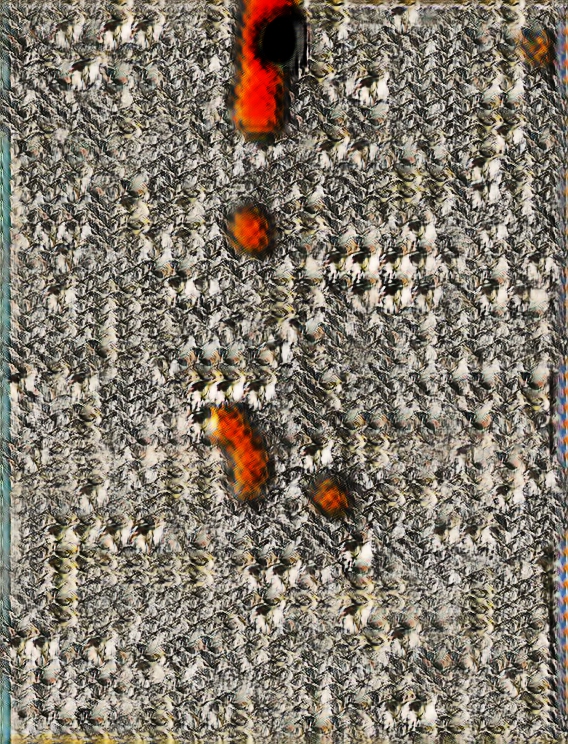}
 \hspace{-5pt}
 \includegraphics[width=0.114\textwidth,height=0.114\textwidth]{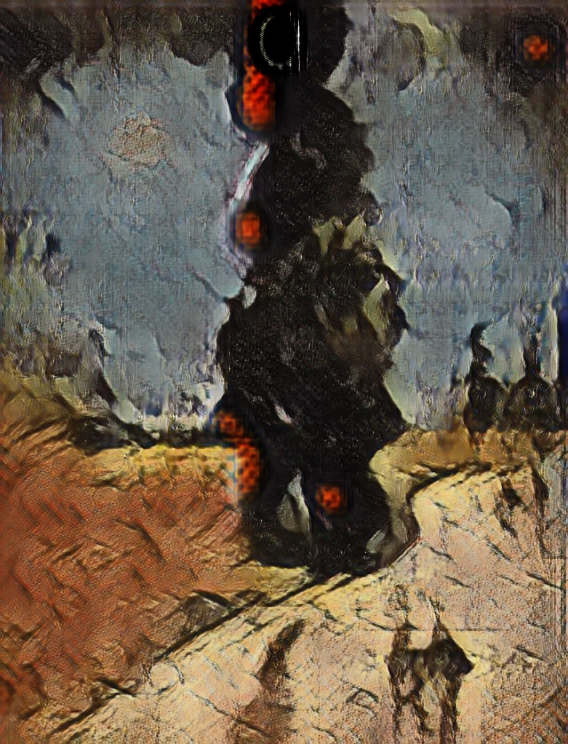}
  \hspace{-4pt}
 \includegraphics[width=0.114\textwidth,height=0.114\textwidth]{fig-sty-drawer/c2}
  \hspace{-5pt}  
  \includegraphics[width=0.114\textwidth,height=0.114\textwidth]{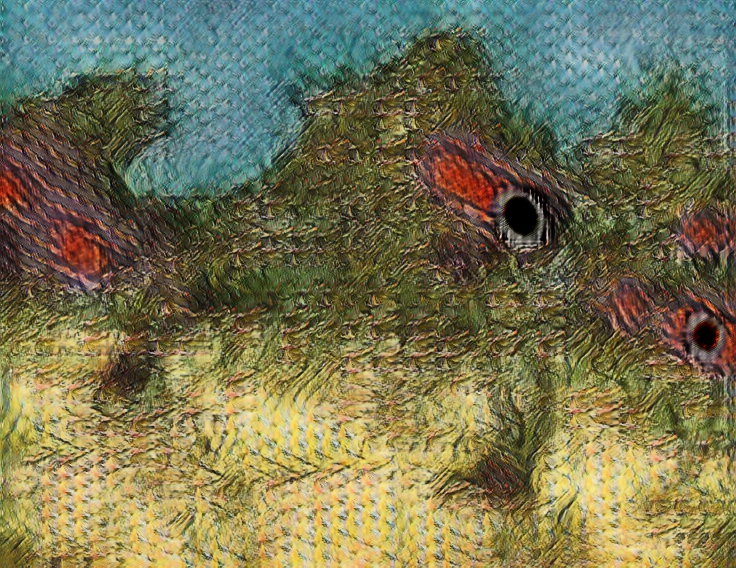}
  \hspace{-5pt}
  \includegraphics[width=0.114\textwidth,height=0.114\textwidth]{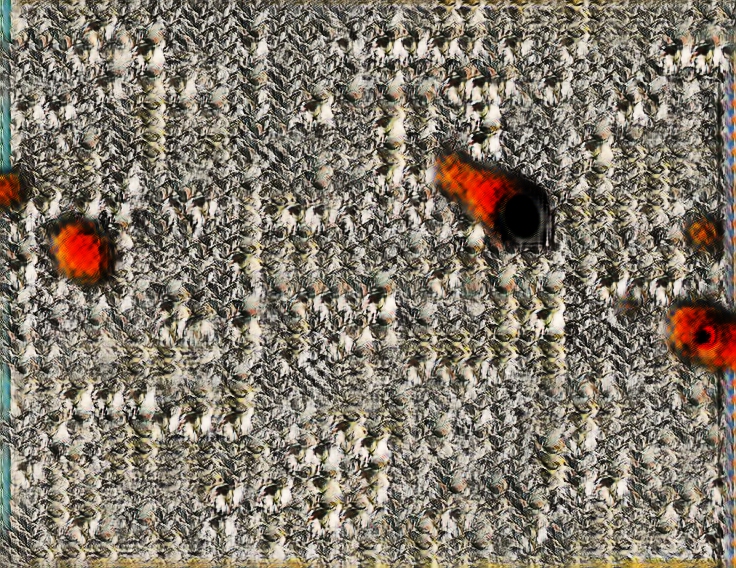}
  \hspace{-5pt}
  \includegraphics[width=0.114\textwidth,height=0.114\textwidth]{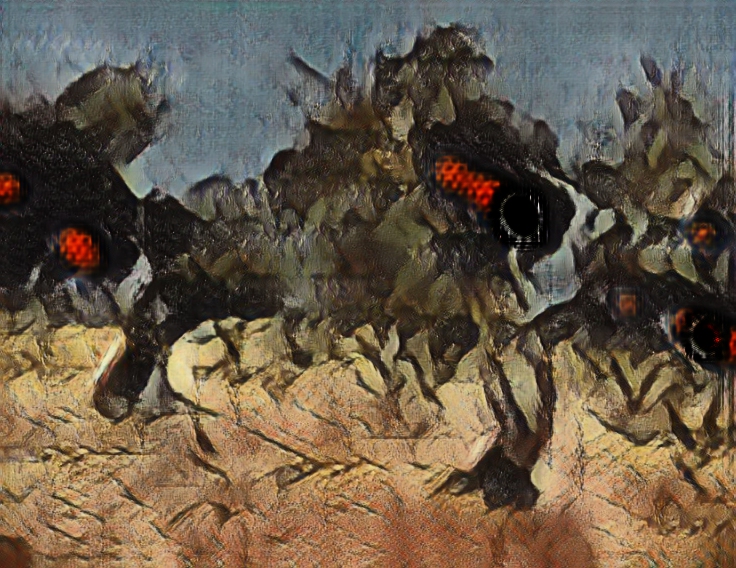}
 \begin{turn}{70} 
 \raisebox{0.04\textwidth }{\makebox[0.05\textwidth]{\centering \scriptsize FUNIT}}
\end{turn}

\vspace{2pt}

 \includegraphics[width=0.114\textwidth,height=0.114\textwidth]{fig-sty-drawer/c1}
  \hspace{-5pt}
 \includegraphics[width=0.114\textwidth,height=0.114\textwidth]{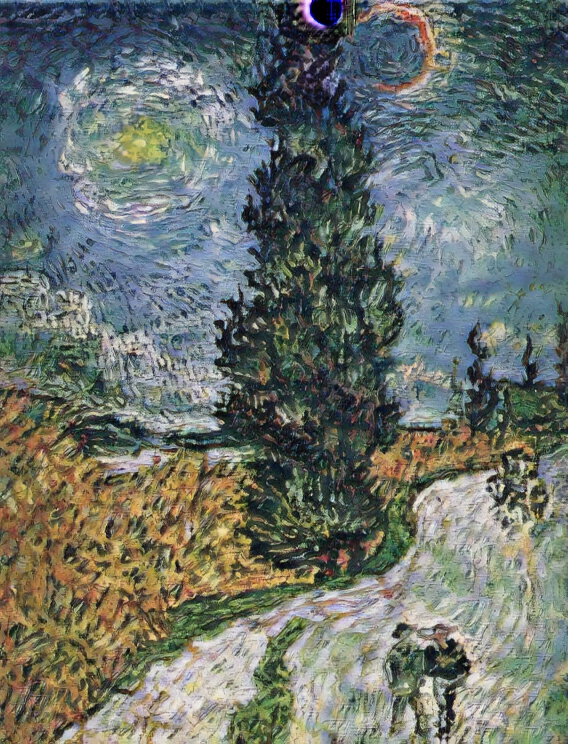}
  \hspace{-5pt}
 \includegraphics[width=0.114\textwidth,height=0.114\textwidth]{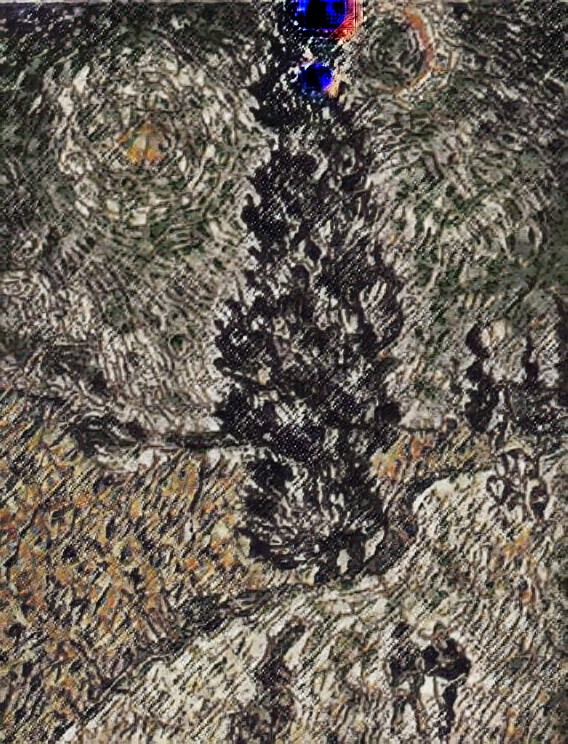}
 \hspace{-5pt}
 \includegraphics[width=0.114\textwidth,height=0.114\textwidth]{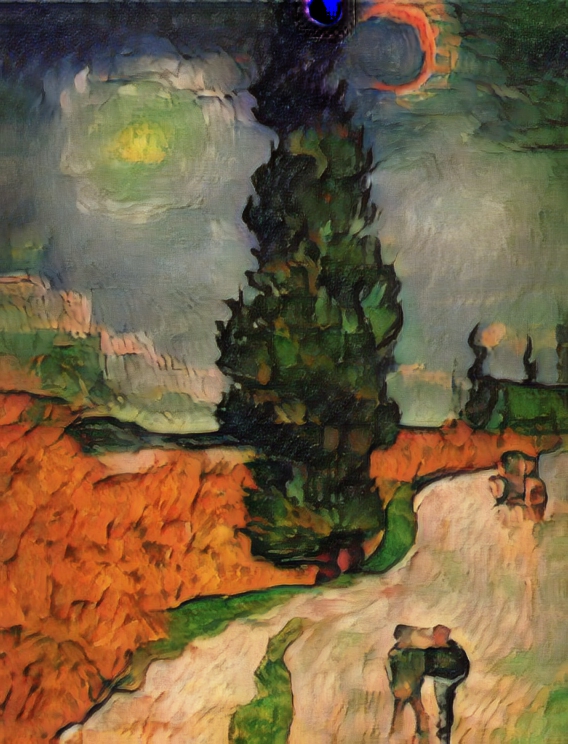}
  \hspace{-4pt}
 \includegraphics[width=0.114\textwidth,height=0.114\textwidth]{fig-sty-drawer/c2}
  \hspace{-5pt}  
  \includegraphics[width=0.114\textwidth,height=0.114\textwidth]{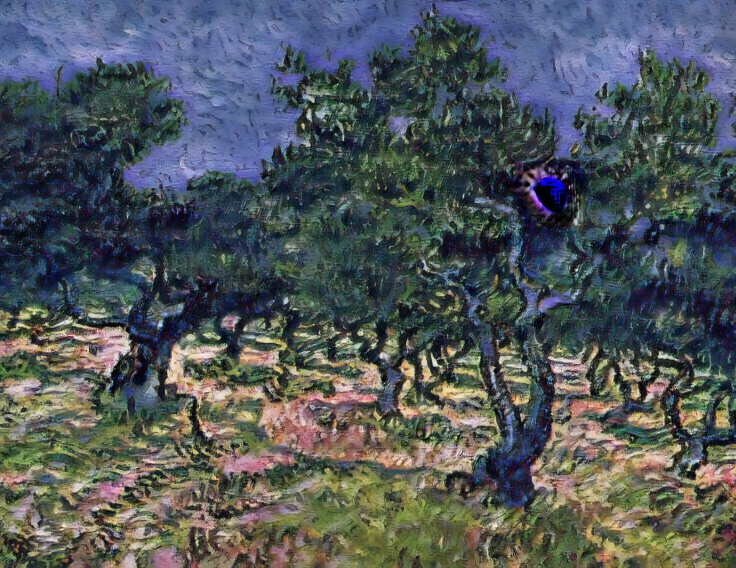}
  \hspace{-5pt}
  \includegraphics[width=0.114\textwidth,height=0.114\textwidth]{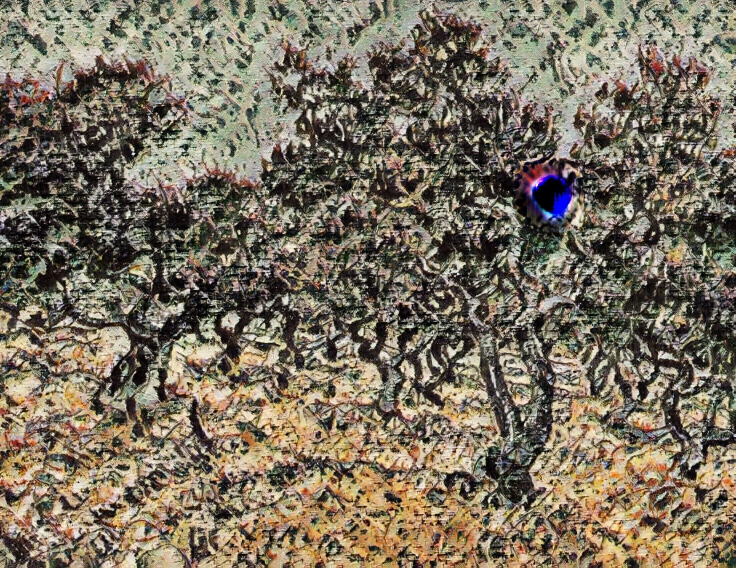}
  \hspace{-5pt}
  \includegraphics[width=0.114\textwidth,height=0.114\textwidth]{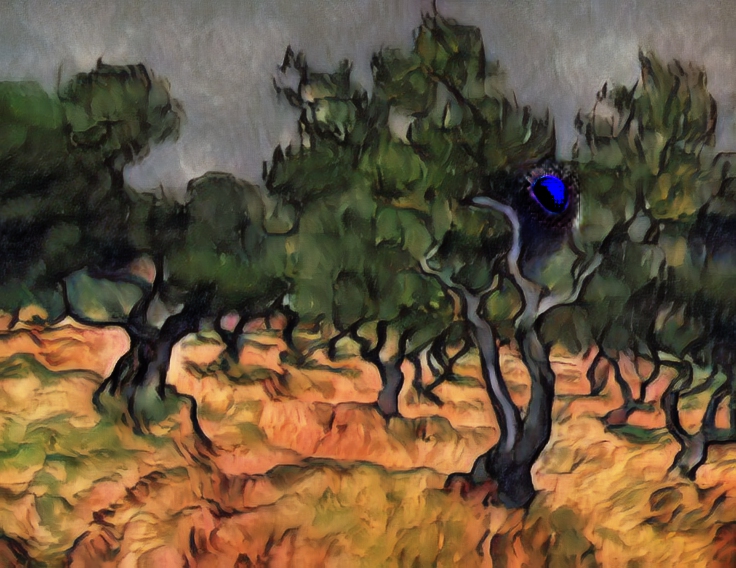}
 \begin{turn}{70} 
 \raisebox{0.04\textwidth }{\makebox[0.05\textwidth]{\centering \scriptsize FUNIT++}}
\end{turn}

 \raisebox{0.005\textwidth }{\makebox[0.99\textwidth][l]{ \scriptsize \hspace{ 5pt}  Van-gogh  \hspace{ 40pt}  Monet \hspace{ 20pt}  Jackson-pollock \hspace{ 20pt}  Cezanne\hspace{ 30pt}  Van-gogh  \hspace{ 30pt}  Monet \hspace{ 30pt}  Jackson-pollock \hspace{ 20pt}  Cezanne}}

\caption{The artistic style images synthesized in high-resolution ($500\times 500$). From left to right: the input images and different artistic styles. Images in the second row are results from the model with the input-output drawer; Images in the third row are results from the model without the input-output drawer; Images in the fourth row are results from FUNIT. FUNIT++ refers to a FUNIT variant trained under a stronger reconstruction constraint. Specifically, we increase the weight of reconstruction loss from 0.1 to 1.0. The red rectangles highlight the regions with artifacts or missing details. Please zoom in for better visualization.}\label{fig:ST}
\vspace{-3mm}
\end{figure*}

\begin{figure}[t]
\centering
\includegraphics[width=1.0\linewidth]{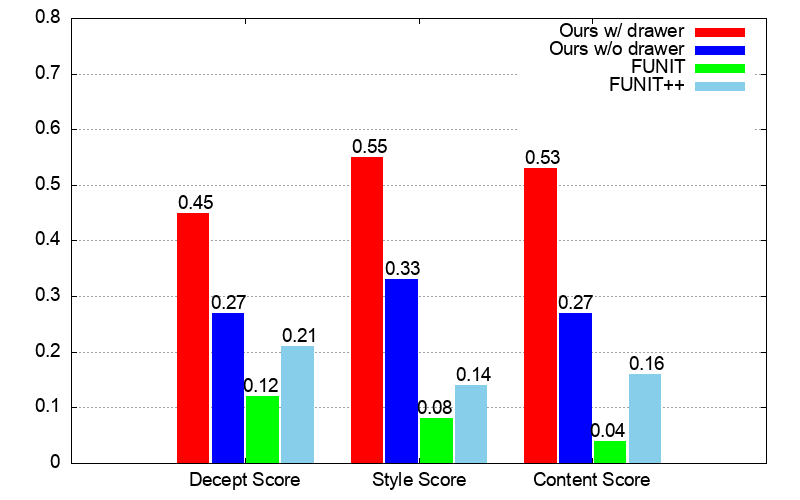}  
\caption{ Quantitative comparison on style transfer task.}\label{fig:style}
\vspace{-0.3cm} 
\end{figure}

\bfsection{Collection Style Transfer} The previous tasks consist of domains that are distinct from each other. To demonstrate the effectiveness of our model. We employ our model for style transfer. We use images from 4 artists and consider each painting style as one domain. This task is challenging because the style images of each artist comprised several distinct styles. It is hard to extract domain-specific features.  We present stylized results of the evaluated methods in Figure \ref{fig:painting}. Our model generates impressive images given input images from different domains. As we can see, the results of season are acceptable but the style translation
of paintings is not so good accompanied by artifacts and
blurriness. Compared to previous tasks, image style
translation needs more variations on texture and color, a
single model might be difficult to simultaneously handle all
styles with large variation. However, our model has the potential for further explorations and extensions.

\section{Discussion}
In this paper, we propose a unified model to handle different issues in image-to-image translation tasks. Our multi-scale discriminator is the driven power to enforce the output to appear like drawn from
the target domain (\eg mappings between different domains). Since there could be
an unlimited number of mappings, we introduce the perceptual self-regularization to encourage perceptual similarity and the input-output drawer to improve the quality of local texture. To verify the effectiveness of each module in the proposed framework, we have performed a further study to compare the performance of our full model with different variants.

\subsection{Effect of the Dilated Convolutional layer in Discriminator}
The dilated discriminator encourages global shape changes with a large receptive field. When combining classifiers at multi-scales with the discriminator for domain-specific feature recognition, this discriminator can provide globally consistent information for the generator training.  To clarify, we compare our model to its variants of discriminator with different numbers of dilated convolutional layers. More dilated convolutional layers indicate a larger receptive field in the discriminator. Because our focus is to learn a mapping for large shape deformation, we evaluate our model on anime to human task. Note that the anime to human task is more challenging than the human to anime task. This is because human face consists of local textures that are missing in anime face. To translate an anime face to human face, the model needs to handle both high-level attributes and local textures. As shown in Figure \ref{fig:dilate}, the FID score of our model with 3 dilated convolutional layers is 24.47, which is lower than other variants with less dilated convolutional layers.

\begin{table*}[t] \setlength{\tabcolsep}{5pt}
\scriptsize 
\RawFloats
\caption{Network architecture for $256\times256$  painting style dataset. Conv(d,k,s) and DeConv(d,k,s) denote the convolutional layer and transposed convolutional layer with d as dimension, k as kernel size
and s as stride. IN is instance normalization. We do not use batch normalization or instance normalization in the discriminator.}	\label{arch}
	\begin{center}
		{\footnotesize
					\renewcommand{\arraystretch}{1.2}
		\begin{tabularx}{\textwidth}{  Y*{3}{|Y} }
			\hlineB{2}
			{\bf Encoder} & {\bf Decoder}& {\bf Discriminator} & {\bf Classifier} \\
			\hline\hline
			Conv(64,4,2), IN, Leaky ReLU &DeConv(1024,4,2), IN, ReLU&\multicolumn{2}{c}{ Conv(64,4,2), Leaky ReLU}  \\
			\hline
			ResNet(64,3,1)&DeConv(512,4,2), IN, ReLU&\multicolumn{2}{c}{ Conv(128,4,2), Leaky ReLU} \\
			\hline			
			Conv(128,4,2), IN, Leaky ReLU& DeConv(256,4,2), IN, ReLU & \multicolumn{2}{c}{ Conv(64,4,2), Leaky ReLU} \\
			\hline			
			ResNet(128,3,1)& DeConv(128,4,2), IN, ReLU &  \multicolumn{2}{c}{ Conv(256,4,2), Leaky ReLU} \\
			\hline
			Conv(256,4,2), IN, Leaky ReLU& DeConv(64,4,2), IN, ReLU&  \multicolumn{2}{c}{ Conv(512,4,2), Leaky ReLU} \\
			\hline
			ResNet(128,3,1)& DeConv(3$\times$ N,3,1), Tanh &  \multicolumn{2}{c}{ Conv(1024,4,2), Leaky ReLU} \\
			\hline			
			Conv(512,4,2), IN, Leaky ReLU& & FC(1024), Leaky ReLU&FC(1024), Leaky ReLU\\
			\hline
			Conv(1024,4,2), IN, Leaky ReLU&&FC(1)&FC(13), Softmax\\
		   \hlineB{2}			
		\end{tabularx}
	}
	\end{center}
\end{table*}

\begin{table}[t!] \setlength{\tabcolsep}{5pt}
\scriptsize 
\RawFloats
	\begin{center}
			\caption{The cost of parameters and model capacity in the process of training. The advantage of the unified methods in retrenching calculating space and time is obvious.}
			\label{complex}
		{\footnotesize
					\renewcommand{\arraystretch}{1.2}
		\begin{tabularx}{\textwidth}{ c *{2}{|Y} }
			\hlineB{2}
			Method &$\#$ Parameters & $\#$ Models with m=10 \\
			\hline\hline
			MUNIT &54.06M& $C_m^2 =\frac{m(m-1)}{2}=45$\\ 
			CycleGAN &52.6M&$C_m^2 =\frac{m(m-1)}{2}=45$\\
			DRIT& 123.42M & $C_m^2 =\frac{m(m-1)}{2}=45$\\
			FUNIT& 46.24M & $1$\\
			StarGAN& 53.28M & $1$\\			
			Ours& 69.74M &  $1$\\
			\hlineB{2}
		\end{tabularx}
	}
	\end{center}
	\vspace{-0.3cm}	
\end{table}

\subsection{Effect of the Input-output Drawer}
To further evaluate the effect of our proposed drawer structure, we perform another experiment on a dataset containing painting collections from multiple artists, including Van-Gogh, Monet, Cezanne, and Gauguin. This task is more challenging than the aforementioned collection style transfer task, since we attempt to generate stylized images in high-resolution. This requires the capacity of the model to preserve the style details such as the color, stroke size, and texture pattern. To demonstrate the advantage of the drawer in preserving local textures, we train one model with the input-output drawer and another model without the input-output drawer for reference. During the training stage, each image is resized to have a smaller edge of 512 pixels while its original aspect ratio remains unchanged. Then a $512\times512$ image patch is randomly extracted as the input that is used to train the model. At the testing stage, the full image with a smaller edge of 512 pixels is fed to the model for style transfer. The results are shown in Figure \ref{fig:ST} and \ref{fig:style}. We observe that the model with the input-output drawer yields the best synthesized style images with well-reconstructed details. While the images created by the model without the drawers contain artifacts in local textures.


\subsection{Effect of the Perceptual  Self-Regularization}
Our model proposes to use a single generator and is trained
with a self-regularization term that enforces perceptual similarity between the output and the
input, together with an adversarial term that enforces the output to appear like drawn from
the target domain. The translated image should preserve both high-level attributes and local texture from the target domain.  In other words, the output and input need to share perceptual similarities, such as the color and the pose,  while allowing changing the domain-specific features. In contrast, the cycle loss seeks for pixel-level consistency. The main issue of the cycle-consistency is that it assumes the translated images contain all the information of the input images in order to reconstruct the input images. This assumption leads to the preservation of source domain features, such as traces of objects and texture, resulting in unrealistic results. Moreover, the cycle-consistency constrains shape changes of source images (\eg results produced by CycleGAN in Figure \ref{fig:Anime}). To verify, we train three model variants: (a) ours w/o $\mathcal{L}_D$ refer to one model trained without the adversarial loss; (b) ours w/ cycle loss stands for one model trained with the cycle loss instead of the perceptual self-Regularization loss; (c) ours w/ perceptual loss represents our full model trained with perceptual self-Regularization loss. Quantitative results of the FID scores at the training stage are reported in Figure \ref{fig:curv}. Our w/o $\mathcal{L}_D$ produces a higher FID score indicating that this model can not learn a mapping between different domains. This proves that the adversarial loss is the driven power to enforce the output to appear like drawn from the target domain. Even though ours w/ cycle loss model achieves a lower FID score, the generated human face contains all the information of the input anime images. However, our full model achieves the best FID score and creates realistic human images.  

\section{Model Complexity Analysis}
We compared the overall model capacity with other
baselines. The number of models and the number of model parameters on the dataset for different $m$ image domains are shown in Table \ref{complex}. 
CycleGAN, DRIT, MUNIT are unsupervised methods, and they
require $C^2_m$ models to learn m image domains, but each of them contains two generators
and two discriminators, while StarGAN and FUNIT only need to train one model to learn all the mappings of $m$ domains. We also report the number of parameters on the face aging dataset,
this dataset contains 10 different age periods, which means $m=10$. Note that our model takes the drawers in the first and last layers in the generator, which brings additional parameters.

\section{Network Structure}
The proposed model consists of one unified generator and discriminator for multi-domain image translation. Table \ref{arch} shows the
detailed network architectures of our model. The discriminator
D is a stack of convolutional layers followed by fully
connected layers, and the classifier C has a similar architecture
and shares all convolutional layers with D. The generator contains one pair of encoder and decoder. Following the Unet, we utilize the lateral connection between the encoder and decoder, which have been shown to produce high-quality results on the image translation task. The encoder is a stack of convolutional layers and the decoder is a stack of transposed convolutional layers. The Architectures for 128 $\times$ 128 images are used in the comparisons with
other baselines on the tasks, human2anime, photo2portrait, animal image translation, and face aging. 256 $\times$ 256 images are used on the task of collection image transfer. 256$\times$256 images are shown in this experiment for better visual effect.

\section{Conclusion}
We have presented a novel GAN model for image-to-image translation across multi-domains. Instead of taking the cycle-consistency constraint, the proposed model exploits a perceptual self-regularization constraint. Specifically, we propose the input-output drawers to preserve domain-specific features. By exploring multi-scale information, our model effectively handles large shape deformation. Extensive experiments demonstrate that the proposed method promotes the unified network to go beyond simple texture transfer and improves shape deformation across multiple domains. The system is capable of performing challenging translations such as from human to anime and from cat to dog. The generated images possess a higher quality compared with the baselines.





\ifCLASSOPTIONcaptionsoff
  \newpage
\fi

{\small
\bibliographystyle{IEEEtran}
\bibliography{egbib}
}

\end{document}